\documentclass[manuscript]{acmart}

\AtBeginDocument{%
  \providecommand\BibTeX{{%
    \normalfont B\kern-0.5em{\scshape i\kern-0.25em b}\kern-0.8em\TeX}}}

\usepackage{amsmath}

\usepackage{multirow}
\usepackage{array}
\newcolumntype{C}[1]{>{\centering\arraybackslash}m{#1}}
\DeclareMathOperator*{\minimize}{minimize}

\begin{document}

\title{A Survey of Adversarial Defences and Robustness in NLP}

\author {
    Shreya Goyal
}
\affiliation{%
  \institution{Robert Bosch Centre for Data Science and AI, Indian Institute of Technology
    Madras}
  \streetaddress{Bhupat and Jyoti Mehta School of Biosciences}
  \city{Chennai}
  \state{Tamil Nadu}
  \country{India}
  \postcode{600036}
}

\author {
Sumanth Doddapaneni
}
\affiliation{%
  \institution{Robert Bosch Centre for Data Science and AI, Indian Institute of Technology
    Madras}
  \streetaddress{Bhupat and Jyoti Mehta School of Biosciences}
  \city{Chennai}
  \state{Tamil Nadu}
  \country{India}
  \postcode{600036}
}

\author {
Mitesh M. Khapra
}
\affiliation{%
  \institution{Robert Bosch Centre for Data Science and AI, Indian Institute of Technology
    Madras}
  \streetaddress{Bhupat and Jyoti Mehta School of Biosciences}
  \city{Chennai}
  \state{Tamil Nadu}
  \country{India}
  \postcode{600036}
}

\author {
    Balaraman Ravindran
}
\affiliation{%
  \institution{Robert Bosch Centre for Data Science and AI, Indian Institute of Technology
    Madras}
  \streetaddress{Bhupat and Jyoti Mehta School of Biosciences}
  \city{Chennai}
  \state{Tamil Nadu}
  \country{India}
  \postcode{600036}
}

\renewcommand{\shortauthors}{Shreya Goyal, et al.}
\newcommand{\NEW}[1]{\textcolor{blue}{#1}}
\newcommand{\sd}[1]{\textcolor{red}{#1}}

\begin{abstract}
 In the past few years, it has become increasingly evident that deep neural networks are not resilient enough to withstand adversarial perturbations in input data, leaving them vulnerable to attack. Various authors have proposed strong adversarial attacks for computer vision and Natural Language Processing (NLP) tasks. As a response, many defense mechanisms have also been proposed to prevent these networks from failing. The significance of defending neural networks against adversarial attacks lies in ensuring that the model's predictions remain unchanged even if the input data is perturbed.
Several methods for adversarial defense in NLP have been proposed, catering to different NLP tasks such as text classification, named entity recognition, and natural language inference. Some of these methods not only defend neural networks against adversarial attacks but also act as a regularization mechanism during training, saving the model from overfitting. This survey aims to review the various methods proposed for adversarial defenses in NLP over the past few years by introducing a novel taxonomy. The survey also highlights the fragility of advanced deep neural networks in NLP and the challenges involved in defending them.
\end{abstract}

\ccsdesc[500]{Computer systems organization~Embedded systems}
\ccsdesc[300]{Computer systems organization~Redundancy}
\ccsdesc{Computer systems organization~Robotics}
\ccsdesc[100]{Networks~Network reliability}

\keywords{Adversarial attacks, Adversarial defenses, Perturbations, NLP}

\maketitle

\section{Introduction}
Recently, there have been significant advancements in the field of Natural Lanaguage Processing (NLP) using deep learning algorithms. In fact, the proposed solutions for NLP have already surpassed human accuracy in some cases \cite{firestone2020performance,lertvittayakumjorn2021explanation, khurana2022natural}. By learning from vast amounts of available data, deep learning has revolutionized the field by providing a representation of language that can be utilized for a range of tasks. NLP involves the manipulation and processing of human language, and deep neural networks have enabled NLP models to learn how to represent language and solve tasks such as text classification, natural language inferencing, sentiment analysis, machine translation, named entity recognition, malware detection, reading comprehension, textual entailment with remarkable accuracy \cite{khurana2022natural, torfi2020natural,  otter2020survey}. 
A typical NLP pipeline using deep neural networks learns word representation in text and contextual details in sentences, and this type of language modeling can be utilized for tasks such as sentence classification, translation, and question answering, using Convolutional Neural Network (CNN) or Recurrent Neural Networks (RNN) based learning models. Finding feature representations for natural language text is a crucial part of this pipeline, and various methods have been proposed, including hand-crafted features and auto-encoded features using recent deep neural networks \cite{torfi2020natural, otter2020survey,lee2022meta}.

Despite the significant progress made, deep neural networks still suffer from a lack of interpretability and operate as a "black box" \cite{buhrmester2021analysis}. Their high performance remains inexplicable, and there is limited understanding of how they function \cite{buhrmester2021analysis,rudin2019we}. Although they can achieve exceptional accuracy and human-like performance, these networks are vulnerable to attacks and are highly sensitive to even the slightest perturbations in inputs, causing them to fail \cite{akhtar2018threat, firestone2020performance}. Recently, there has been a growing number of proposed adversarial attacks for deep neural networks in computer vision and NLP, raising concerns about the robustness of these high-performing models \cite{zhang2020adversarial, yuan2019adversarial}. Adversarial attacks can pose a significant security threat to applications such as malware and spam detection, as well as biometrics. In NLP, these attacks can take various forms, including substitution, insertion, deletion, and swapping of words/characters in a sentence or finding a neighboring word embedding to introduce perturbations in the input \cite{zhang2020adversarial}. 

There are two main types of adversarial attacks, black box and white box, based on the attacker's access to the model's parameters \cite{zhang2020adversarial}. These attacks can be further categorized based on their design granularity, including character level, word level, sentence level, and multi-level attacks \cite{zhang2020adversarial,wang2019towards,huq2020adversarial}. Adversaries are generated by perturbing the input text using techniques such as insertion, deletion, flipping, swapping of characters or words, or paraphrasing the sentence in a way that preserves its original meaning but changes the wording. In white box attacks, the attacker has access to the model's parameters and modifies the word embeddings of input text using gradient-based schemes. In contrast, black box attacks do not have access to the model's parameters and generate a replica of the model by repeatedly querying the input and output. Once the parameters are acquired, they train a substitute model with perturbed data and attack it \cite{zhang2020adversarial,wang2019towards,huq2020adversarial,yuan2019adversarial, qiu2022adversarial}. Generating perturbations in textual data is more challenging than in images due to the discrete nature of the data \cite{zhang2020adversarial}. The quality of adversarial examples generated for text data is determined by two factors, namely, the naturalness of the adversarial examples and the efficiency to generate these examples \cite{li2023efficiently}. Some researchers have succeeded in detecting perturbations in text using simpler techniques like spell check and adversarial training \cite{pruthi2019combating}, while others who used word-level attacks failed to efficiently generate adversarial examples due to the high-dimensional search space \cite{zang2019word}. Thus, efficiently generating adversarial attacks in NLP poses unique challenges. Despite the difficulty, stronger and imperceptible adversarial attacks have been proposed, which pose a significant threat to the security of deep neural networks \cite{boucher2022bad, ballet2019imperceptible}. Consequently, several defense mechanisms have been proposed in recent years to counter adversarial attacks in NLP, and the considerable amount of work in adversarial defenses has provided good competition to the novel adversarial attack algorithms, substantially improving the robustness of existing deep learning models.

Adversarial defense strategies in NLP can be broadly classified into three categories: adversarial training-based, perturbation control-based, and certification-based methods. The majority of the work in this field employs an adversarial training approach and techniques are further subdivided based on the generation of adversarial instances or noise in the defense pipeline. These methods include data augmentation-based adversarial training, adversarial training as a regularization technique, Generative Adversarial Network (GAN)-based adversarial training, Virtual Adversarial Training (VAT), and Human-In-The-Loop (HITL) approaches. Perturbation control-based methods are also categorized into perturbation identification and correction and perturbation direction control. Certification-based techniques fall under the third category and provide certificates of robustness against adversarial attacks. A few methods do not fit into any of the aforementioned categories and are classified as miscellaneous. In the subsequent section, the objectives of this survey paper are emphasized, distinguishing it from previous surveys.
\vspace{-5mm}

\subsection{Goals of this survey paper}
In this article, we reviewed numerous methods of adversarial defenses in NLP, proposed in recent years. The key goals of this survey are as listed below:
\vspace{-5mm}

\begin{itemize}
\item Providing a comprehensive review of adversarial defense schemes in NLP by covering schemes for different NLP tasks and bringing the attention of the community to this emerging area. 
    \item Proposing novel taxonomy for adversarial defense methods in NLP for various tasks. 
    \item Accentuating  the importance of defense methods for adversarial attacks and as a regularization scheme in deep neural networks. 
    \item Paving the path for future work in this area by highlighting the open issues. 
\end{itemize}

Numerous survey papers have been published in the past discussing adversarial attacks on deep neural networks in both computer vision and NLP. For example, in \cite{xu2020adversarial}, the authors conducted a comprehensive survey of adversarial attacks on deep neural networks for images, text, and graphs, proposing a novel taxonomy to categorize a wide range of methods. Similarly, \cite{akhtar2018threat} proposed a taxonomy for different adversarial attacks and defenses on various computer vision algorithms, including image classification, image segmentation, object detection, robotic vision, and visual question answering. In addition, \cite{ozdag2018adversarial} presented a brief survey on general adversarial attack methods in deep learning, while \cite{kong2021survey} briefly reviewed attack and defense methods in images and text data. Furthermore, \cite{chakraborty2018adversarial} and \cite{bhambri2019survey} discussed adversarial attacks and defenses for various computer vision algorithms. 

In contrast to previous research, \cite{zhang2021survey} proposed a novel taxonomy for universal adversarial attacks, which includes universal perturbations for image classifiers, and briefly discussed attacks on text and audio classification models. While the previously discussed survey papers focused primarily on adversarial attacks on images and briefly discussed attack algorithms in NLP, \cite{zhang2020adversarial,wang2019towards,huq2020adversarial,qiu2022adversarial} extensively reviewed adversarial attack algorithms for various NLP tasks while briefly discussing some defense methods. However, the importance of adversarial defense algorithms is self-evident, given the large amount of work in this area in recent years. Therefore, this survey paper aims to address this gap and is different from previous survey papers in this area by focusing exclusively on adversarial defense methods in NLP. This paper proposes a detailed and novel taxonomy for adversarial defense mechanisms in NLP, emphasizes the importance of defense methods, and discusses open issues in this area while presenting future work to the community.

In this paper, Section \ref{sec:adversarialattacks} discusses adversarial attacks in deep learning and categorizes adversarial attacks in NLP. Section \ref{sec:taxonomy} presents a novel taxonomy for adversarial defense methods in NLP. Section \ref{sec:advtrain} provides a detailed description of adversarial training-based defenses in NLP, along with sub-categories. Section \ref{sec:perturbationcontrol} discusses perturbation control-based adversarial defense methods. Section \ref{sec:certification} outlines certification-based adversarial defenses. Section \ref{sec:misc} describes various other adversarial defenses that do not fit into the previous categorization. Section \ref{sec:metrics} discusses different metrics used to evaluate defense mechanisms. Section \ref{sec:datasets} describes the datasets and frameworks proposed for training and evaluating adversarial defense methods. Section \ref{sec:futurework} provides suggestions for future research in adversarial defenses for NLP, and finally, Section \ref{sec:conclusion} concludes the paper.

\section{A general overview of adversarial attacks}
\label{sec:adversarialattacks}
An adversarial attack is a deliberate attempt to corrupt a deep neural network's functionality by introducing distorted inputs that cause the model to fail. These perturbations are designed to be subtle enough to evade human detection but effective enough to deceive a neural network. For instance, image classification models have been subjected to experiments with various input perturbations, including the addition of noise, the adjustment of pixels, the use of patches, the addition of watermarks, and so on, which can go unnoticed by humans. In contrast, adversarial attacks in NLP involve multiple proposed perturbations at the character, word, or sentence level through deletion, insertion, swapping, flipping, use of synonyms, concatenation with characters or words, insertion of numeric or alphanumeric characters, etc. However, it is more challenging to generate adversarial perturbations for text data than image data because altering a character or word in a sentence is more perceptible to humans. Moreover, creating imperceptible adversarial attacks is difficult in NLP since perturbations in textual data could result in less natural input data \cite{li2023efficiently}. Adversarial attacks are classified into two categories based on motivation: targeted attacks and non-targeted attacks. Targeted attacks aim to misclassify inputs to a specific class, while non-targeted attacks aim to push the classifier boundary to cause the model to misclassify inputs. Based on access to the model's parameters, adversarial attacks are classified as white-box and black-box attacks. In this section, we briefly review the state-of-the-art adversarial attacks for NLP tasks algorithms.

\subsection{Adversarial attacks in deep learning}
The goal of an adversarial attack is to generate such perturbations for input $\mathbf{x}$ belonging to class $\mathbf{C_1}$, such that, $\mathbf{x}$ is wrongly classified to class $\mathbf{C_2}$ with a high confidence value. For a multi-class classification algorithm, for $k$ input classes $i = C_1, C_2,... ,C_k$, the perturbed input is $\mathbf{x^{'}}$ and $\mathbf{f_i}$ is the discriminant function which defines the classification boundaries, where $\mathbf{x}$ belongs to class $\mathbf{C_i}$, and $\mathbf{C_{target}}$ is the target class after the attack, then:
\begin{equation}
\label{eq:inequality} 
    {f_{target}(\mathbf{x'})}> { f_i(\mathbf{x'})} 
\end{equation}

Hence, adversarial attacks can be formally defined as an optimization problem for $\mathbf{x}$ as:

\begin{equation}
\begin{gathered}
\minimize_{x}  (||x'-x||) \\
{ subject \  to\ \max_{i \neq target}} \{{f_i(\mathbf{x'})}\}-   {f_{target}(\mathbf{x'})} \leq 0
\end{gathered}
\label{two}
\end{equation}
The inequality defined in \ref{eq:inequality} represents the goal of any adversarial attack which pushes the perturbed input $\mathbf{x'}$ to a desired target class, rather than its actual class. Hence Equation \ref{two} defines the adversarial attack as an optimization problem,  where the goal is to minimize the perturbation magnitude to make the perturbations less perceptible and make sure it gets classified to the target class, $\mathbf{C_{target}}$ \cite{ozdag2018adversarial, ren2020adversarial, xu2020adversarial,ma2021understanding}.


\subsection{Adversarial attacks in NLP}
In the past few years, numerous methods for adversarial attacks have been introduced, which are specifically designed for NLP tasks. It is important to note that adversarial examples in computer vision cannot be directly applied to text as they are fundamentally different. Therefore, several attack methods that modify the text data while maintaining imperceptibility to humans have been proposed in literature. Typically, these methods alter the text data at the word, character, or sentence levels. The following section presents some of these attack methods in NLP.

\textbf{Character level adversarial attacks}
Character-level attacks perturb the input sequences at a character level. These operations include insertion, deletion, and swapping of characters in a given input sequence. Despite the fact, these attacks are quite effective, they can easily be detected with a spell-checker mechanism. One of the techniques used in character-level attacks is adding natural and synthetic noise to the inputs \cite{belinkov2018synthetic}. For natural noise authors collected natural spelling mistakes and used them to replace words in inputs. For synthetic noise, they swap or randomized characters (except the peripheral) and replace a character with its neighboring character on the keyboard. Adding punctuation marks, and increasing or removing the space between characters is another technique to add synthetic noise in the text inputs. For example, in DeepWordBug \cite{DBLP:conf/sp/GaoLSQ18}, in black-box setting they use a two-step process as they don’t have access to the gradients, parameters, or structure of the model. The first step involves finding the most important words in the sentence which would be the target words to perturb. In the second stage, perturbations are added to these select words by the above-mentioned operations. Edit distance is further used in order to keep track of the readability of the generated sentences. Another example proposed is, TextBugger \cite{li2018textbugger} in both black-box and white-box settings where the white-box attack is a two-step process.
The first process involves finding the most important word with the help of the jacobian matrix ($\mathcal{J}_\mathcal{F}(x)$) defined as $\mathcal{J}_\mathcal{F}(x)=\frac{\partial \mathcal{F}(x)}{\partial x} = $
$  \left[  \frac{\partial \mathcal{F}_i(x)}{\partial x_i} \right]_{i \in 1...N, j \in 1...K} $, where $x_i$ is the $i^{th}$ word of the input text, $N$ is the total number of words in the input text, and $\mathcal{F}$ is the classifier, and later use $5$ different options to add bugs. These $5$ include insert, delete, swap, substitution with visually similar words, and substitution with a semantically similar words. In the black-box setting they propose a 3-step process where first the most important sentence is identified, then they find the important words to generate 5 bugs and select the optimal from that. The best adversary is chosen based on how optimal they are for reducing accuracy.
Along the same line, \cite{hosseini2017deceiving} has shown that just by adding extra "." (period), spaces between words, "Perspective" API created by Google gave lesser toxicity scores for the words perturbed in this fashion.


\textbf{Word level adversarial attacks }
Word level attacks perturb the whole word instead of a few characters. Common operations include insertion, deletion and replacement.
Word level attacks can be classified into Gradient-based and Importance based and replacement-based strategies on the basis of the perturbation schemes used:
\begin{itemize}
    \item In gradient-based methods, the gradient is monitored for every input perturbation. Whenever the probability of classification is reversed that particular perturbation is chosen. This is inspired by the Fast Gradient Sign Method (FGSM)
    \cite{goodfellow2015explaining} used for adversarial attacks in computer vision models. If the classification probability changes the class then the perturbation is considered effective. Another way of using gradient based method is to find the important words using FGSM and then employ insertion, deletion and replacement strategies on top of them \cite{samanta2017crafting}. \cite{DBLP:conf/ijcai/0002LSBLS18} used a similar approach where they created adversaries by backpropagating for the cost gradients. 
    \item  In importance-based methods it is believed that words with the highest or lowest attention scores play an important role in predictions of self-attention models. Hence these are chosen as the possible vulnerable words. These words are greedily perturbed until the attack is successful. One of the methods ``Textfooler" \cite{jin2019bert} uses a similar strategy where important words are greedily replaced with synonyms until the classification label changes. Another work in this direction \cite{ivankay2022fooling} proposed TextExplanationFooler algorithm, which designed word importance-based attacks for explanation models in text classification problems. Working in a black box attack setting proposed attack attempted to alter outputs of widely used explanation methods while not changing the predictions of the classifier. 
    \item In replacement-based methods, words are randomly replaced with semantically and syntactically similar words. Here the replacement for words is obtained by using word vectors like GloVe \cite{DBLP:conf/emnlp/PenningtonSM14} or thought vectors. \cite{Kuleshov2018AdversarialEF} used thought vectors to map sentences to vectors and replaced one word from it's nearest neighbors which had best effect on the objective function. \cite{DBLP:conf/emnlp/AlzantotSEHSC18} used GloVe vectors to randomly replace words that fit in context of sentence.
\end{itemize}

\textbf{Sentence level adversarial attacks }
These attacks can be considered as manipulation of a group of words together instead of individual words in the sentence. Moreover, these attacks are more flexible, as a perturbed sentence can be inserted anywhere in the input, as long as it is grammatically correct. These attack strategies are commonly used in tasks such as Natural Language Inferencing, Question-Answering, Neural Machine Translation, Reading Comprehension, text classification. 
For sentence-level attacks novel techniques such as ADDSENT, ADDANY are introduced in literature in recent years with variants such as ADDONESENT, ADDCOMMON \cite{jia-liang-2017-adversarial, wang-bansal-2018-robust}. Some of the sentence based attacks are created such that they don't affect the original label of the input and used as a concatenation in the original text. In these cases, the correct behavior of the model is to retain the original output and the attack is successful if the model changes the output/label.
In another set of methods, GAN based sentence level adversaries are created which are grammatically correct and semantically close to the input text \cite{zhao2018generating}.  Another example ``AdvGen" \cite{cheng-etal-2019-robust} is introduced which is an example of gradient based white-box method and used in neural machine translation models. They used greedy search guided with the training loss to create the adversarial examples while retaining semantic meaning. Another work in this direction, \cite{DBLP:conf/naacl/IyyerWGZ18} proposed syntactically controlled paraphrase networks (SCPNS) for adversarial example generation where they used encoder-decoder network to generate examples with a particular syntactic structure. 

\textbf{Multi-level adversarial attacks}
Multi-level attack schemes consist of a mixture of some of the methods discussed above. These attacks are used to make the inputs more imperceptible to humans and to have a higher success rate. Hence, computationally more intensive and more complicated
techniques such as FGSM have been used to create adversarial examples. In one such method, they create hot training phrases and hot sample phrases. In this method, the training phrases focus on what and where to insert, modify or delete by finding hot sample phrases in white and black box settings where deviation score is used to find the importance of the words \cite{10.5555/3304222.3304355}. 
Another example used "HotFlip" \cite{ebrahimi-etal-2018-hotflip} which is a character level white-box attack  swapping characters based on the gradient computation. Similar to many other techniques, TextBugger \cite{li2018textbugger} tries to find the most important word to perturb using a Jacobian matrix in a white box setting. The important words after identification are used for creating adversaries by inserting, deleting and swapping along with Reinforcement Learning methods  with an encoder-decoder framework. 

\section{Taxonomy of adversarial defenses}
\label{sec:taxonomy}

\begin{figure}[h!]
    \centering
    \includegraphics[scale=0.38]{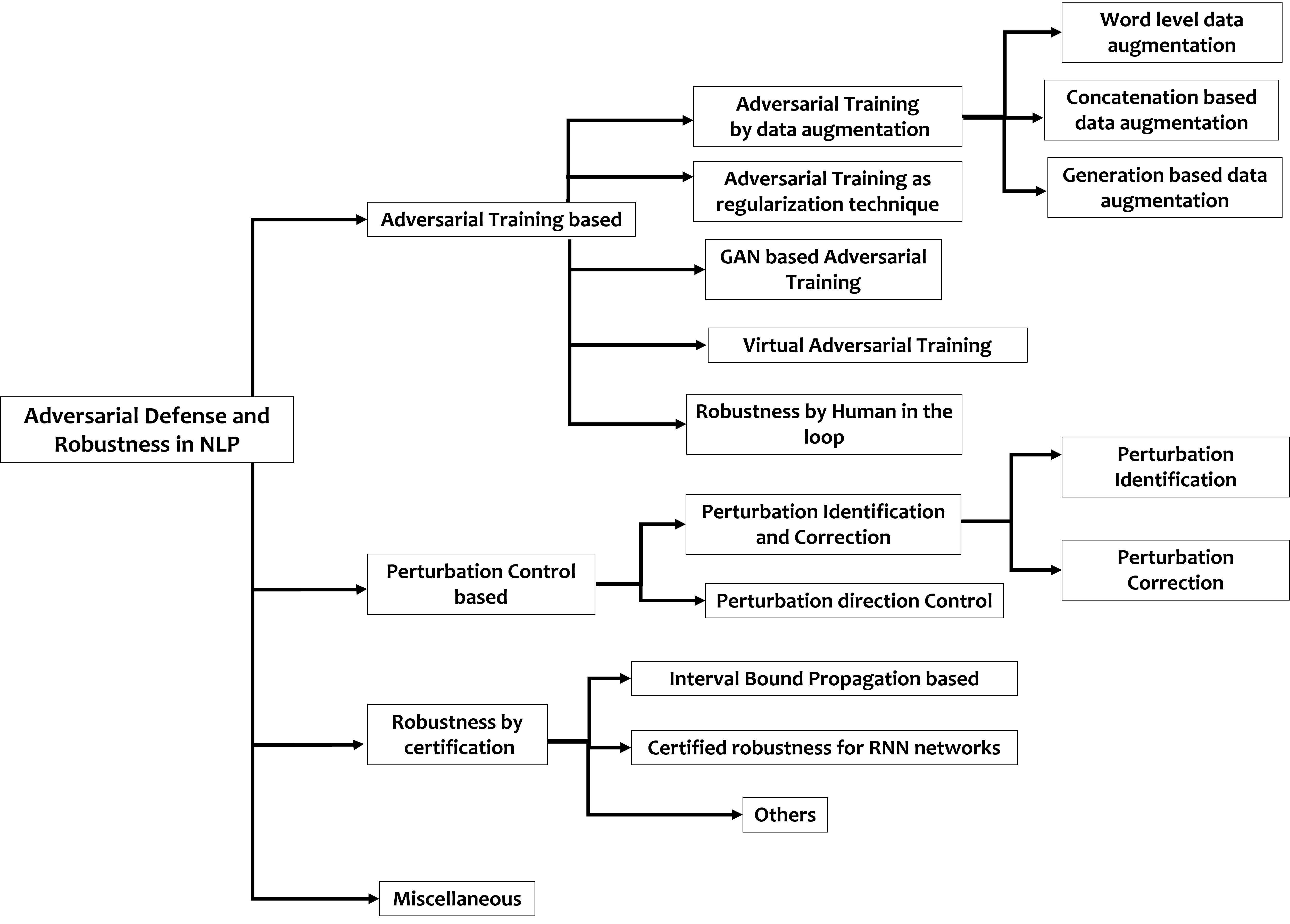}
    \caption{Taxonomy of adversarial defense methods in natural language processing}
    \label{fig:taxonomy}
\end{figure}

In this section, we will discuss the different types of defense methods used to protect deep learning models from adversarial attacks. We will also highlight some of the recent research works that have shown promise in this area. Adversarial defense strategies are methods used to prevent deep neural networks from failing due to adversarial attacks. These methods aim to increase the robustness of neural networks by training them in an environment that simulates adversarial attacks, or by adding mechanisms to detect and handle adversarial inputs. Another approach to increase robustness is to create a perturbation-resistant region around the input space. Therefore, in NLP, there are three main strategies for designing adversarial defense methods: creating a similar environment during neural network training, identifying malicious inputs during training and correcting them using specialized methods, and certifying the robustness of the input region for the network.

The defense methods in NLP discussed in this paper are divided into three main categories: adversarial training based methods, perturbation control based methods, and certification based methods, along with some miscellaneous approaches. Methods that do not fall under the first three categories are included in the miscellaneous category. The first set of techniques falls under the category of (i) \textit{Adversarial Training}, which serves as a defense mechanism against adversarial attacks. The various subcategories of methods included in this group are: Adversarial training through data augmentation, Adversarial training as a regularization technique, Adversarial training using Generative Adversarial Networks (GANs), Virtual adversarial training, and Robustness through human intervention. The second set of methods is based on (ii) \textit{perturbation control}, and includes two subcategories: Perturbation identification and correction, and Perturbation direction control. The third set of methods follows approaches that provide (iii) \textit{certification} of the model's robustness against adversarial attacks. The last set, (iv) \textit{miscellaneous}, comprises a combination of various methods that do not fit into any of the aforementioned categories. Figure \ref{fig:taxonomy} illustrates the detailed taxonomy proposed in this survey article for adversarial defense techniques in NLP.

\section{Adversarial Training based defenses}
\label{sec:advtrain}
Adversarial training was first introduced in the work proposed in \cite{goodfellow2014explaining}. It is a method of defending against adversarial attacks by introducing adversarial examples in the training data. The strength of adversarial examples decides the final robustness and generalization achieved by the model. Adversarial training is further divided into sub-groups on the basis of the strategies used for augmenting the data such as word or character level modification, model-based generation of adversarial examples or adversarial inputs generated by concatenation in the original data. 
While some of the methods performed adversarial training by generating a set of adversarial examples and inserting them in the training dataset, other methods used adversarial training as regularizer by introducing perturbation within the network. In this section, some of the work in literature will be highlighted which proposes to use of data augmentation for generating adversarial examples for adversarial training.

\subsection{Adversarial training by data augmentation}
Adversarial defense methods often use adversarial training as a basic technique for defending against adversarial attacks. This method involves generating a set of adversarial data using a perturbation scheme and incorporating it into the model training process. Many techniques for generating adversarial examples using data augmentation have been proposed in the literature. These techniques involve identifying the most important words, characters, or other parts of the input text that affect the output the most and manipulating the input data by flipping, inserting, deleting, or swapping those parts of the sentence. Concatenating a piece of text in the input by finding the most appropriate position is another strategy used in the literature . There are also automatic methods for generating adversarial examples for adversarial training, some of which will be discussed in the following sections.

\subsubsection{Word level data augmentation}
There are methods in literature that propose to modify words or word embeddings in the input text for generating adversarial examples to augment data. In this line the work proposed in Textbugger \cite{li2018textbugger} presents adversarial training by data augmentation for text classification in both white box and black box settings by generating utility-preserving adversarial examples. In the white box, important words for perturbations are found using Jacobian of classifier and then optimal perturbations are found by searching the nearest neighbor space in Word2Vec embeddings. In black box setting, data is augmented by finding important words and sentences which contribute to the output the most and manipulating them. The work in \cite{cheng2020advaug} proposed novel data augmentation technique for adversarial training in machine translation task, which reinforces the model to virtual data points around the observed examples in training data. They proposed vicinity distribution for adversarial space (space of adversarial examples centered around each training example), and sampled virtual adversarial samples from it using interpolated embeddings of existing training samples. In the work \cite{li2020adversarial} adversarial perturbations are applied on word embedding layer of a CNN for text classification task to make the classification model robust towards the worst perturbations. Another work \cite{yang2021adversarial} proposed a new method PQAT which perturbs the embedding matrix rather than the word vector for machine reading comprehension task. Two additional independent embedding spaces for paragraph-question(PQ) are used, to give additional context for the same word used with different roles in paragraphs and questions. During training P-Q embeddings are added to the original vector keeping the context from passage and question separate. 

In another work, \cite{zhang2020deep}, the authors proposed  continuous bag-of-word (CBOW) embedding based perturbations for generating human imperceptible adversaries for text classification task. Embedding space generated by CBOW is used for predicting the perturbation direction and tries to preserve the meaning of the sentence by placing a constraint on the perturbation direction. Authors generated adversarial examples without altering the semantic meaning of the sentence and used these examples in adversarial training. In the same line, \cite{liu2020joint} proposed a solution to the Out-Of-Vocabulary (OOV) words problem faced by conventional character level defense methods leading to a poor performance of models. They proposed adversarial stability training to overcome these challenges. Stability training is a technique which makes the output of the neural networks significantly robust, while maintaining the original performance \cite{zheng2016improving}. Proposed Adversarial Stability Training (AST) is used with character level embeddings, to overcome OOV problems and adversarial sentences are generated by perturbing each word, using character level embeddings representation to overcome the distribution problem. Another work in the same direction, \cite{hsieh2019robustness} created  adversarial examples using several schemes such as, random word replacement, synonym replacement, finding weak spots in the input strings with greedy approach, by constraining the embeddings within $L_1$ distance, replacing the word on the basis of attention score (high attention score word to low score word) and demonstrate their results on sentiment analysis, textual entailment, and machine translation tasks. They analyzed the robustness of RNNs, transformers and BERT-based models and demonstrated that self-attentive models are more robust than RNNs. Work proposed in \cite{bvelohlavek2017using} evaluated adversarial training with adversarial examples for eight datasets in NLP targeted for several purposes such as question answering, reasoning, detection, sentiment analysis, and language detection using language models such as LSTM and GRU. They used combinations of dropout and adversarial example for evaluation. Another work \cite{staliunaite2021improving} proposed data augmentation for adversarial training for increasing the robustness of causal reasoning task. They proposed data augmentation by synonym substitution and by filtering out casually linked clauses in the larger dataset and used generative language models to generate distractor sentences as potential adversarial examples. 
 To improve the conventional adversarial training methods \cite{wang2021towards} proposed to use gradient based approach for ranking the important words in the training dataset and distilBERT similarity score for finding similarity between two-word embeddings for a faster and low-resource requirement-based adversarial training. Also they propose to use a percentage of training data for generating adversarial examples instead of converting all the training data for cheaper adversarial training. 

In \cite{ebrahimi2018adversarial}, authors have proposed a white box based defense method by generating adversarial examples with flip, swap, insert and delete at character level and used gradient based optimization method to rank the examples. The work \cite{zhang2020generating} proposed Metropolis-Hastings (MH) sampling \cite{chib1995understanding} based adversarial example generator for text classification. Three level word operations, replacement, insertion and deletion are performed with MH sampled words in a black box setting, while the gradient of the loss function is inserted in the pre-selection function of MH in case of white box attack. For the text classification task, in the work \cite{dong2020leveraging} adversarial training is used for cross lingual text classification and robustness enhancement by training it on English data and using the model to predict on non-English unlabelled data. The predicted outputs are used as adversarial examples for adversarial training.
 In the same line, \cite{ren2019generating}, authors have proposed a greedy algorithm probability weighted word saliency for adversary generation for the text classification task. Adversarial examples were generated with word synonym replacement and named entities with other named entities using WordNet, picking up the words which cause maximum change in text classification probability. In the work \cite{zang2019word}, black box adversary generation is proposed which uses sememe (minimum semantic unit in linguistics) based word substitution that is more sememes per sense mean fewer substitute words that share the same sememes can be found, which negatively affects the adversarial attack success rate. Later they used Particle swarm optimization \cite{kennedy1995particle} based algorithm to search the optimal adversarial example. In literature, frameworks and APIs have also been proposed which provides a complete platform to the user with various kinds of attacks to generate adversarial examples to be used for adversarial training for defense. In this line, the work \cite{morris2020textattack} presented a python framework for attack generation and defending the model, which augments the data using a word embedding, word swap, thesaurus word swap, homoglyph character substitution, etc. They also provide a set of constraints on the generated perturbations to keep them indistinguishable from the original such as, grammar check, POS tags etc. Along the same lines, \cite{wallace2020imitation}, generated imitation models for Google API of machine translation and generated adversarial samples using techniques such as flipping, replacing malicious nonsense, substituting phrases that cause incorrect translation and attacking original model in a black box manner. Later they used those adversarial examples to train the imitation model and transfer the examples to the victim model. In this work the authors aim towards, finding vulnerability in the victim model to make it more robust, by having the victim model output a different high-accuracy translation. 
Apart from word level perturbations, there are some sentence level perturbations proposed to generate adversarial examples, which are discussed in coming section. These adversarial examples are hand crafted and generated by identifying vulnerability of model, and further used with adversarial training to defend model from such attacks. However, some of papers proposed model based generation of adversarial examples are also discussed in coming sections.

\subsubsection{Concatenation based data augmentation}
Another approach to generating adversarial examples for data augmentation involves using concatenation-based strategies and automatic generation of adversaries with language models. For instance, \cite{jia2017adversarial} developed concatenative adversarial perturbations, AddSent, to generate adversarial examples for reading comprehension systems. This method involves concatenating grammatically correct sentences to end of a paragraph, which look like questions or other arbitrary sentences. They also replaced some answers with fake answers that have the same part of speech type and category. Similarly, \cite{wang2018robust} proposed AddSentDiverse, which generates adversarial examples with significantly higher variance by varying placement of perturbations. They also expanded set of fake answers used in AddSent and demonstrated the limitations of the original AddSent method.

\subsubsection{Generation based data augmentation}
Another approach to improving the robustness of machine learning models is through generation-based adversarial examples. For example, \cite{kang2018adventure} proposed using knowledge-guided rules and a seq2seq model to generate new hypotheses from a given premise for textual entailment models. In a similar vein, \cite{han2020adversarial} proposed a defense strategy using adversarial training, using a seq2seq model to generate perturbations for structure prediction tasks that involved predicting POS tags, parse-trees, and noun phrases. In \cite{xu2021grey}, the authors proposed a grey box adversarial attack for sentiment analysis that uses a generator model for data augmentation. Adversarial training was conducted by augmenting data using a static copy mask mechanism in the generator, and counter-fitted word embeddings and label smoothing methods were used to better capture lexical relations and preserve the labels of adversaries. \cite{mrkvsic2016counter} introduced the idea of injecting antonymy and synonymy constraints into vector space representations. In the same line, \cite{wang2020cat} proposed a new method of adversarial example generation by controlled adversarial text generation where they aimed to perturb input for a given task by changing other controllable attributes of the dataset. For example, in the case of sentiment analysis task for product reviews, product category becomes a controllable attribute that cannot change the sentiment of a review. Their pretraining module consists of an encoder-decoder architecture, which is used to teach the model to copy the input sentence S, assuming that it has the controllable attribute ($a$) in the sentence. The decoder module is updated to generate a sentence containing attribute $a'\neq a$. In the optimizing module the subspace of all $a'$ is checked by computing the cross-entropy loss to find the highest perturbation. In similar lines of developing robust NLP model, the work proposed in \cite{la2022king} studied a group of linguistic rules to demonstrate local semantic robustness within a sentence and generated variations in input text using predefined template with fixed label. These templates adhere to the linguistic rules discussed in the paper, where incoming variations will not be able to change the output label and further used them for training robust sentiment analysis models. In the direction of paraphrasing-based adversarial instance generation, \cite{iyyer2018adversarial} proposed syntactically controlled paraphrase networks (SCPNs), which generate a paraphrase of the given sentence with the desired syntax in a controlled manner. SCPNs use a bidirectional LSTM  and a two-layer LSTM augmented with soft attention over the encoded states based encoder-decoder architecture, utilizing a paraphrase pair and a target syntax tree as the inputs. Using the adversarial instances in adversarial training, they evaluated the proposed method on sentiment classification and textual entailment applications. 

\begin{table}[h!]
    \centering
    \begin{tabular}{||l|c|p{30mm}|p{50mm}|p{20mm}||}
    \hline
     \textbf{Strategy}    &  \textbf{Work} & \textbf{Granularity} & \textbf{Application} & \textbf{Threat Model}\\
     \hline
       &  \cite{li2018textbugger}  & Words \& Sentences & Sentiment Analysis& White Box, Black Box\\ \cline{2-5}
    
        & \cite{cheng2020advaug} &  Sentence level & Machine Translation & Black Box\\ \cline{2-5}
        & \cite{li2020adversarial} & Words & Text Classification & White Box\\ \cline{2-5}
        & \cite{yang2021adversarial} & Embedding Matrix & Reading Comprehension & White Box \\ \cline{2-5}
        
        & \cite{mrkvsic2016counter} & Word embedding & Dialogue state tracking & White Box \\ \cline{2-5}
        & \cite{zhang2020deep} & Word embeddings  & Text Classification & White Box \\ \cline{2-5}
        & \cite{liu2020joint} & Character Embeddings & Text Classification & White Box\\ \cline{2-5}
    \multirow{2}{*} {Word}     & \cite{hsieh2019robustness} & Word & Sentiment Analysis, Textual Entailment, Machine Translation & White Box\\ \cline{2-5}
        & \cite{bvelohlavek2017using} & Word and Character embeddings  & Question Answering, Reasoning, Sentiment Analysis, Language detection & White Box\\ \cline{2-5}
        & \cite{staliunaite2021improving} & Word level & Causal Relation classification & White Box \\ \cline{2-5}
        & \cite{ebrahimi2018adversarial} & Character level & Machine translation & White Box\\ \cline{2-5}
        & \cite{zhang2020generating} & Word level & Text classification & Black box\\ \cline{2-5}
        & \cite{ren2019generating} & Word level & Text Classification & White Box\\ \cline{2-5}
        & \cite{zang2019word} &  Word level & Text classification & Black Box\\ \cline{2-5}
        
        & \cite{wallace2020imitation} & Phrase level  & Machine Translation & Black Box\\ \cline{2-5}
        & \cite{dong2020leveraging}  & Word level & Document \& Intent  classification & White Box \\ \cline{2-5}
        
        \hline
       Concatenation &\cite{jia2017adversarial}& Sentence & Reading comprehension & White Box \\ \cline{2-5}
        &\cite{wang2018robust} & Sentence  & Reading comprehension & White Box\\ \cline{2-5}
        \hline
        &\cite{kang2018adventure}&  Sentence generation & Textual Entailment & White Box\\ \cline{2-5}
        &  \cite{han2020adversarial} & Sentence generation & Predicting POS tags, parse trees, NP & Black Box\\ \cline{2-5}
      Generative  & \cite{xu2021grey} & Sentence generation  & Sentiment Analysis & Grey Box\\ \cline{2-5}
        
        & \cite{wang2020cat} & Sentence generation& Sentiment Analysis & White Box\\ \cline{2-5}
        & \cite{la2022king} & Sentence generation & Sentiment Analysis & White Box \\ \cline{2-5}
        & \cite{iyyer2018adversarial} & Sentence generation & Sentiment classification, Textual entailment & White Box \\
        \hline
    \end{tabular}
    \caption{Summary of the adversarial training methods by data augmentation }
    \label{tab:advtraining}
\end{table}

Table \ref{tab:advtraining} shows the summary of all the adversarial training methods with data augmentation. It summarizes the strategies used in the literature for perturbation generation, along with the granularity of the perturbations used, demonstrated NLP applications, and the kind of the threat model they are defending. Many of the word-based methods for data augmentation involve operations like flipping, swapping, inserting, deleting, and synonym substitution with words to modify the original inputs and generate adversarial examples for use with adversarial training. Some methods use these operations with characters in the sentences instead of whole words. These adversarial perturbation generation methods aim to maintain the naturalness of the input text so that the perturbation is imperceptible to humans. Therefore, some of these methods use word embeddings for perturbations instead of words from the input text. Another set of methods in the literature use concatenation operations within input sentences by identifying the appropriate position for concatenation, and these adversarial examples are used in adversarial training. In addition to manual perturbations and adversarial example generation, another direction of methods uses generative models to create adversarial examples or perturbations. Along with the data augmentation methods, adversarial training methods introduce perturbations within the training loss functions, which are discussed in the next section.

\subsection{Adversarial training as regularization technique}
In this section, another defense method based on adversarial training for NLP is discussed. In this style of adversarial training, the input perturbations are incorporated as a part of model training, instead of training it with adversarial examples. In the work \cite{goodfellow2014explaining}, authors proposed adding perturbations in input as a regularizer in the loss function. The modified optimization function based on the fast gradient sign method after adding perturbations is defined as:
\begin{gather*}
    L_{adv}(x_l, \theta)= D[q(y|x_l), p(y|x_l+ r_{adv}, \theta)]\\ 
    r_{adv}=\underset{r;||r||_2 \leq \epsilon}{\operatorname{argmax}}  D[q(y|x_l), p(y|x_l+ r, \theta)]
\end{gather*}

Where, $L_{adv}$ is the adversarial loss term, $r_{adv}$ is the adversarial perturbation, $x_l$ is the labeled input data, $D$ is the non-negative divergence measurement function between two probability distributions, $\epsilon$ is the upper bound on the perturbations, $q(y|x_l)$ is the unknown true distribution of the output label. This loss function is designed to approximate the true distribution $q(y|x_l)$ by a parametric model $p(y|x_l, \theta)$ which is robust against adversarial attack to the input data $x$. Adversarial training using perturbations in the loss function has been shown to be a successful defense strategy as it generates adversarial examples that are difficult to create manually by exploiting flaws in the optimization function to improve model generalization. In the field of NLP, several techniques have been proposed to introduce perturbations during training or modify the loss function. Here, we describe some of the methods that have been proposed to introduce perturbations during training.
  
 In the work \cite{miyato2016adversarial}, the authors included perturbations at each step of training and tried to minimize the loss function for text classification tasks. In another work \cite{yasunaga2017robust}, authors proposed adversarial training for POS tagging, by using character level embeddings with BiLSTM models, where word-level embeddings are generated by concatenating character level embeddings. Perturbations are added to the input at the character level embeddings in the direction which maximises the classifier loss function while training the model. Another method \cite{wu2017adversarial}, introduced adversarial noise at embedding (concatenation of word and characters) level for the task of relation extraction within the multi-instance multi-label learning framework. They proposed a joint model to entity recognition and relation extraction using adversarial training as a regularization scheme where worst case perturbations are added to maximize the training loss. Along the same line, \cite{wang2019improving} authors improved the neural language modeling by using adversarial training as a regularization technique. They injected an adversarial perturbation on the word embedding vectors in the softmax layer of the language models while training the model. Authors have suggested new loss functions and diverse neural network optimization methods to train a model adversarially and improve its robustness.
In this direction, the work \cite{minervini2018adversarially} proposed adversarial training for natural language inferencing, by reducing the adversarial example generation problem to combinatorial optimization problem. They proposed a continuous inconsistency loss function that measures the degree to which a set of examples can cause a model to fail. By maximizing the inconsistency loss and constraining the perplexity of the generated sentences, adversarial examples are generated, posing it as an optimization problem.
In the same direction, \cite{kariyappa2019improving} proposed defense mechanism, diversity training, for transfer based attacks for the ensemble of models. They proposed gradient alignment loss (GAL) which is used as regularizer to train an ensemble of diverse models with misaligned loss gradients.  Another work \cite{dong2020towards} proposed a novel Adversarial Sparse Convex Combination (ASSC) method to leverage regularization term for introducing perturbations. They modeled the word substitution attack space as a convex hull of word vectors and further proposed ASSC-defense for using these perturbations in adversarial training.

There are methods in the literature that proposed novel regularization techniques for enhancing the robustness of a language model. In this line of work \cite{wang2020infobert} proposed InfoBERT for increasing the robustness of BERT based models by analyzing language models from information-theoretic perspective. They presented two mutual information based adversarial regularizers for adversarial training. Information Bottleneck regularizer which extracts minimal features for downstream tasks and removes noisy and vulnerable information for potential adversarial attacks. Anchored Feature regularizer extracts strong local features which are not vulnerable while aligning local features with global features to increase the robustness. 
Another work \cite{zhu2019freelb} proposed FreeLB, to use K-step PGD \cite{athalye2018obfuscated} for generating adversaries in adversarial training and used multiple PGD iterations. In contrast with k-step PGD and freeAT \cite{shafahi2019adversarial}  methods, used multiple PGD iterations to create adversaries simultaneously accumulates the “free” parameter gradient. In this method while optimizing the objective function, it replaces the batch of input with K times large batch which include perturbations along with inputs.
Improving the performance of this work \cite{li2021searching} proposed FreeLB++, by extending the search region to a larger $l_2$-norm and increasing the number of search steps at the same time since FreeLB has a narrow search space. They also bench-marked various defense methods in literature under standard constraints and settings to have a fair comparison of these methods. Table \ref{tab:advreg} describes the summary of adversarial training as regularization technique. 
In the coming section GAN based adversarial defense methods are discussed, where GAN is used as a generator and discriminator model to build more robust models against adversarial attacks. 

\begin{table}[h!]
    \centering
    \begin{tabular}{||c|c|c||}
    \hline
    \textbf{Work} & \textbf{NLP task} & \textbf{Granularity}\\
    \hline
     \cite{miyato2016adversarial}    & Text Classification & Word embedding\\
     \hline
       \cite{yasunaga2017robust}  & POS Tagging & Character embeddings\\
       \hline
       \cite{wu2017adversarial} & Relation Extraction & Word and character embeddings\\
       \hline
       \cite{wang2019improving}  & Machine translation, language modeling & Word embeddings\\
       \hline
       
\cite{minervini2018adversarially} & Natural Language Inferencing & Sentence Embedding \\
\hline
\cite{dong2020towards} & Sentiment Analysis and Natural Language Inferencing & Word embeddings\\
\hline
\cite{wang2020infobert} & Question Answering, Natural Language Inferencing & BERT embeddings\\
\hline
\cite{zhu2019freelb} & Natural Language Inferencing, Textual Entailment & Word embeddings\\
\hline
\cite{li2021searching} & Sentiment analysis, Text Classification & Word embeddings\\
\hline
       
    \end{tabular}
    \caption{Summary of adversarial training as regularization technique}
    \label{tab:advreg}
\end{table}

\subsection{GAN based adversarial training}
Using GAN for adversarial training is another approach that has been used in literature for defending models from adversarial attacks. In this approach a generator and discriminator are trained together in adversarial manner, where the generator is primarily used for generating adversarial examples. The discriminator is responsible for discriminating between the clean data and the adversarial training samples to make model more robust towards adversarial attacks. 
In this line, the work proposed in \cite{kang2018adventure} uses adversarial training withva GANs for the textual entailment task. The generator and discriminator are trained in an end-to-end manner, where generator (seq2seq) is trained for generating adversarial examples using external knowledge or handwritten rules. Discriminator is trained in the same manner to learn the textual entailment for the generated samples. In another work, \cite{ren2020generating}, authors used a conditional variational autoencoder which generates the adversarial examples for text classification task. They further used a discriminator and GAN training framework for adversarial training and to make sure the generated adversaries are consistent with the real-world data. In the work \cite{liu2017adversarial} authors used multi-task learning for domain adaptation. They proposed a model which has separate units to discriminate between shared patterns and task specific patterns using GAN for creating adversarial examples and includes this loss in the final loss for optimization. They demonstrated the improved performance over $16$ datasets and the learned parameters in both shared and task-specific parts of the network. The work in \cite{xu2019lexicalat} proposed a new adversarial training approach that mimics a GAN. The generator is used to create adversarial examples with the help of lexical knowledge base where a classifier is used to score the generated adversarial example. The model is trained in a reinforcement learning fashion due to discrete-time generation of the generator model and the score from the classifier is used as the reward for the generator. The generator generates examples by replacing words in the input sentence with synonyms, neighboring words, and superior words. 
In another work, \cite{coavoux2018privacy} try to defend against an attacker who tries to take the encoded information to reconstruct the original input text. In the adversarial training-based defense strategy they used GAN style training with 2 components in which the original one predicts the class of a given sentence and a binary classifier predicts the privacy element. The training style involves making the output prediction as correct as possible while at the same time creating more complex examples for the privacy element classifier. Table \ref{tab:advgan} describes the summary of GAN based adversarial training methods, where each row describes the NLP task for which defense is designed and the specific strategy used in the proposed work. 


\begin{table}[h!]
    \centering
    \begin{tabular}{||c|c|c||}
    \hline
    \textbf{Work} & \textbf{NLP Task} &  \textbf{Strategy} \\
    \hline
        \cite{kang2018adventure}  & Textual Entailment  & Using external knowledge\\
         \hline
      \cite{ren2020generating}   &  Text classification & Using Conditional Variational Autoencoder (VAE)\\
       \hline
      \cite{liu2017adversarial} & Text classification & Adversarial shared-private model\\
       \hline
      \cite{xu2019lexicalat}  & Sentiment Analysis & Using lexical knowledge base\\
       \hline
      \cite{coavoux2018privacy}  & Sentiment Analysis and Topic classification & Privacy measurement of neural representations\\
      \hline
    \end{tabular}
    \caption{Summary of GAN based adversarial training methods}
    \label{tab:advgan}
\end{table}

\subsection{Virtual Adversarial Training (VAT)}
Virtual Adversarial Training (VAT) is another variant of adversarial training-based defense methods first proposed by \cite{miyato2018virtual}. VAT is found to be a very efficient method in the case of semi-supervised learning methods because it defines the adversarial direction without label information. In contrast to Adversarial training, VAT doesn't need full label information for generating perturbation. The intuition behind VAT is to add perturbation $'r'$ to input $x$ such that the divergence of their output space is maximum. Hence, training is done in a way to minimize the divergence term after adding the perturbed input to make the model robust against adversarial attacks.  The modified loss function for virtual adversarial training is defined as:
\begin{gather*}
    LDS(x_*, \theta)= D[p(y|x_*, \hat{\theta}), p(y|x_*+ r_{adv}, \theta)]\\ 
    r_{adv}=\underset{r;||r||_2 \leq \epsilon}{\operatorname{argmax}} D[p(y|x_*, \hat{\theta}), p(y|x_*+ r_{adv})]\\
    R_{adv}(D_l, D_{ul}, \theta)= \frac{1}{N_l+ N_{ul}} \sum_{x_* \in D_l, D_{ul}} LDS(x_*, \theta) 
\end{gather*}

Where, $x_*$ are the ``virtual" labels which are probabilistically generated and $r_{adv}$ are virtual adversarial perturbations. Here, $x_*$ is kept in the place of $x$, since label information is not available for all the input data. $LDS$ is the local smoothness term for the input data point $x$ and $R_{adv}$ is the final regularization term. 

The work proposed in \cite{miyato2016adversarial} extended the notion of virtual adversarial training and adversarial training for text classification and sequence models proposing this technique as a regularization method. For defending the models, they introduced perturbations in word embeddings of the text inputs, while minimizing the KL divergence of VAT. 
In another work \cite{pereira2021targeted}, authors proposed a VAT method by performing adversarial steps on those examples which are predicted as wrong by the model and then regularise the model for this target direction in contrast with general adversarial training methods where the perturbation is done for all examples with variation from the gold label. In a targeted training manner, they try to steer the examples to a particular label $y_t$ and presented a comparison with human-annotated data along with other adversarial training algorithms.
In the same direction, the work \cite{liu2020adversarial} authors proposed a novel adversarial robust model ``Adversarial training for large neural LangUage Models(ALUM)" for  defending BERT-based pretraining language models. It is a general model for adversarial training in pretraining and fine-tuning which regularizes the training objective by applying perturbations in the embedding space that maximizes the adversarial loss. The model is regularized using VAT. Experimenting with different word embeddings using VAT, \cite{zhang2020deep} extended the adversarial training regularization for semi-supervised tasks. They used continuous bag of words (CBOW) model for generating word embeddings and restricted perturbation directions for creating adversaries. Targeting specifically sequence labelling tasks in NLP, \cite{chen2020seqvat} proposed VAT for sequence labelling task combining CRF, making sequence labelling task more robust. They use CNN layer for extracting character and word embeddings, LSTM for sequence encoding, and CRF decoder layer to incorporate the probabilities of label transition. 
Introducing more variations to VAT, \cite{li2020tavat} proposed Token aware virtual adversarial training. In contrast with conventional virtual adversarial training, TAVAT generated token aware perturbations instead of random perturbations to avoid unnecessary noise and take important information carried by tokens into consideration. Table \ref{tab:VAT} describes the summary of Virtual Adversarial Training (VAT) based methods along with the specified NLP task for their design and granularity of the perturbation.

 \begin{table}[h!]
     \centering
     \begin{tabular}{||c|c|c||}
     \hline
        \textbf{Work}  & \textbf{NLP task} &  \textbf{Granularity}   \\
        \hline
        \cite{miyato2016adversarial}  & Text classification \& Sequence modeling & Word embeddings \\
        \hline
        \cite{pereira2021targeted} & Natural Language Inferencing (NLI) &   Word embeddings \\
        \hline
        \cite{liu2020adversarial} & Question answering, NLI,  Named Entity Recognition(NER) & BERT embeddings \\
        \hline
        \cite{zhang2020deep} & Sentiment classification & Word embeddings \\
        \hline
         \cite{chen2020seqvat} & Chunking, NER, slot filling & Character and word embeddings \\
         \hline
         \cite{li2020tavat} & NER, NLI, Textual Entailment, text classification & Token level\\
         \hline
     \end{tabular}
     \caption{Summary of Virtual Adversarial Training (VAT) based defense methods}
     \label{tab:VAT}
 \end{table} 

Along with the several variants of adversarial training methods, there are a few schemes that utilizes Human-In-The-Loop (HITL) framework, where human-level intervention is considered for adversarial training. Some of these methods are discussed in the next section.

\subsection{Robustness by human in the loop}
Human-In-The-Loop (HITL) is an idea of leveraging human intervention while training models or defending them against adversarial attacks. The scheme is extensively used in various fields of artificial intelligence. It takes advantage of both human and machine intelligence, for labeling the data, and training and validation of models. While it is proven to be an efficient scheme in other areas of artificial intelligence, several authors have tried to use HITL for developing algorithms for adversarial defenses. 

The work \cite{zang2019word} proposed a sememe-based word substitution method to generate perturbations. Apart from using particle swarm based optimization algorithm to search perturbation for data augmentation, they manually selected $692$ valid adversarial samples for adversarial training to further boost the performance. Also, authors in \cite{nie2019adversarial,williams2020anlizing}, created dataset of adversarial examples, ANLI for natural language inferencing task by crowd-sourcing. Adversarial examples are written and verified by human annotators in $3$ stages in loop, while getting them tested from high performing NLI models. They also presented error analysis and discuss annotation scheme and data collection process of ANLI. In the work \cite{dinan2019build} authors built a model for offensive language detection in dialogues using human and models in loop. They trained BERT based model on Wikipedia Toxic Comments dataset, and asked crowd workers for marking the messages as offensive if they are wrongly marked safe by the system. This process is performed in multiple iterations to build a robust system. 
Work in \cite{wallace2019trick} presented a defense method using Human-In-The-Loop by proposing human-computer hybrid approach for evaluating the models. They presented a human verification of the question-answering system, where human annotators authored adversarial examples to break a model-based QA system but still answerable by humans. This process is targeted towards building a robust question-answering system by inserting human-authored adversarial examples. Table \ref{tab:advhitl} depicts summary of the Human-In-The-Loop (HITL) based adversarial training methods in NLP defense. Each row in table depicts the NLP application for which the defense method is designed and granularity at which human interaction is proposed.  
Adversarial training-based defense methods are evidently the most popular way to build robust models against adversarial attacks in NLP. It requires the generation of adversarial examples using several techniques and using them for training the model. However, in contrast with training with adversarial examples, there are methods in the literature that are proposed for the detection of perturbations, their correction, and controlling directions of those perturbed instances. Some of these methods are discussed in the next section.

\begin{table}[h]
    \centering
    \begin{tabular}{||c|c|p{50mm}||}
    \hline
    \textbf{Work} & \textbf{NLP Tasks} & \textbf{Human interaction granularity }\\
    \hline
      \cite{zang2019word}   &  Sentiment Classification, NLI & Manual selection of word substitution based perturbations  \\
      \hline
        \cite{nie2019adversarial}  & NLI & Crowd Sourced and tested with models \\
        \hline
        \cite{williams2020anlizing} & NLI & Crowd Sourced and tested with models\\
        \hline
        \cite{dinan2019build} & Offensive language detection & BERT with crowd workers\\
        \hline
        \cite{wallace2019trick} & Question Answering & Human annotators and verification\\
        \hline
    \end{tabular}
    \caption{Summary of the Human In The Loop (HITL) based adversarial training methods}
    \label{tab:advhitl}
\end{table}

\vspace{-10mm}

\section{Perturbation control based defenses}
\label{sec:perturbationcontrol}
The adversarial defense methods proposed in previous sections, use data augmentation schemes, perturbation generation within training for supervised and semi-supervised tasks, adversaries monitored by human and models in loop, defending the model in a generating and discriminating manner. However, all these schemes do not incorporate the idea of interpretable perturbations or reconstruction of generated perturbations. In literature, there are schemes that control direction of perturbations to make the perturbations more meaningful, indistinguishable and re-constructive and further use them in training. Also, another set of method try to identify the perturbed inputs and correct them to make the models more robust. In this line, the following sections describe methods which have been proposed in this direction. 

\subsection{Perturbation identification and correction }
In literature perturbation, control-based method tries to identify malicious inputs after an attack and correct them. Some of the methods only identify these inputs and filter the training data accordingly. While the other category of methods also corrects these inputs after their identification using methods such as spell checker or rule-based. Some of these methods are described below. 

\subsubsection{Perturbation Identification}
Perturbations related to word modifications which included insertion, deletion, substitution or swapping of words are identified in several ways. One of those methods is proposed in \cite{wangnatural}, where the authors proposed defense mechanism, against synonym substitution, calling it “Synonym Encoding Method”(SEM). They essentially clustered all the synonyms in embedding space with their euclidean distances and then encoder is layered before input to train the model. Encoder is responsible for identifying all the synonym substitution-based attacks in the model and maps all the synonyms to a unique encoding without adding extra data for training. Improving on this work \cite{yang2022robust} proposed a robust adversarial training method called Fast Triplet Metric Learning (FTML). This method tries to cluster the similar embedding and pushes dissimilar embedding where SEM works directly with input texts and establishes no relation with non synonym clusters. FTML forces each word with a similar meaning to have the same representation in the feature space and pushes away the word with a different meaning. The method incorporates a word level triplet loss which tries to minimize the distance between a word with its corresponding group of synonyms and maximizes the distance with its non-synonym group. In another work in this direction, \cite{zhou2020defense} proposed Dirichlet Neighborhood Ensemble (DNE), a randomized smoothing method for training a robust model to defend substitution-based attacks. DNE forms virtual sentences by sampling embedding vectors for each word in an input sentence from a convex hull spanned by the word and its synonyms, and it augments them with the training data, (mixing the embedding of the original word in the input sentence with its synonyms). The work in the same line \cite{bao2021defending} introduced frequency-aware randomization framework Anomaly Detection with Frequency Aware Randomization (ADFAR) for defense against adversarial word substitution. They add an extra module to detect perturbation in the sentences and apply ADFAR only on sentences that are identified as adversarial. This module is added to the language model and use a multi-task learning procedure. They demonstrated  that ADFAR works better than other defenses on 4 datasets - MR, SST-2, IMDb, MNLI. The work \cite{wang2021randomized} proposed an adversarial defense scheme by perturbation detection for synonym substitution attacks. They proposed a novel method, namely, Randomized Substitution and Vote (RS\&V). The proposed method calls an input text ``adversarial example", by randomly substituting some of its words by their synonyms and checking the consistency of the highest voted label for all perturbed examples. If the label is found to be inconsistent with the original label then it is considered as an adversarial input. Working in a similar direction, \cite{swenor2022using} proposed to use of random perturbations to defend sentiment analysis models. They inserted random perturbations to the multiple copies of a randomly selected sentence from the reviews. These perturbations included synonym substitution, dropping of a word or a spell check with correction if necessary. All these perturbed copies of the sentences are put together with the original review and a majority vote is taken for each sentence by the sentiment classifier. By majority-based classification, the models are taken back to their original performance accuracy before the attack.
In the direction of perturbation identification, \cite{xie2022identifying} proposed an extensive dataset, TCAB, for attack detection and labeling with over $1.5$ million instances of adversaries. To construct this dataset, authors used 6 datasets for text classification with 3 target classifiers and 12 different attacks from textAttack and openAttack and presented a benchmark for these attacks. They proposed an attack identifier/labeler, using three features of the input such as text properties including contextual embeddings, length of the input, token case, punctuation, non ASCII characters. Authors use language model's properties as another feature that identifies the structure of the language such as phrasing or ungrammatical input text. They used target model's properties as another feature that captured the changes in model's gradients, activation or saliency because of malicious inputs and trained a classifier with them. Utilizing the TCAB dataset and the text, language, and classifier features, the detection of the perturbation and labeling the type of attack is proposed. Another work \cite{wang2021textfirewall} proposed a novel method called TextFirewall identification of adversarial inputs. They used word importance to quantify the importance of a word in the input sentence to the final classification of the model. The impact of each word in an input sentence is calculated to detect the perturbation, by summing the scores of each model, and then the model's output is compared with original ground truth to identify the perturbed input. In a similar direction, \cite{yoo2022detection} proposed an adversarial robustness method by proposing feature density estimation-based perturbation detection. In contrast with the available frequency-based likelihood estimation, they utilized the probability density of sentences. They proposed a method of Robust density estimation (RDE), which fits the probability density estimation model on the features obtained from a pre-trained model like BERT. Dimensionality reduction is applied to the parameters of these features to avoid the curse of dimensionality. In addition, they released a benchmark for word-level adversarial detection using 4 NLP models with four different datasets for text classification. In addition to the discussed work in perturbation identification, some of the methods in literature also attempt to correct these perturbed input data after their detection. Some of the perturbation correction methods are discussed in next section. 
\subsubsection{Perturbation correction}
Adversarial inputs are also required to be corrected after their identification to retain the training data. Some of the methods in this direction of work attempted to identify perturbations related to character-level modifications in the input text. In this direction, the work \cite{sakaguchi2017robsut} authors proposed a semi-character level recurrent neural network (ScRNN), which act as a spell checker by recognizing words. ScRNN has an architecture similar to standard RNN and takes semi-character vector as input and predicts a correct word at each time step by applying three types of noises: jumble, delete, and insert.
As an extension of the above work, in \cite{pruthi2019combating} authors try to combat misspellings by using a word classifier before the actual classifier of a task. They propose ScRNN with backoff, to overcome limitations of ScRNN \cite{sakaguchi2017robsut}, and propose three backoff techniques if the word classifier predicts it as unknown (UNK). As a backing-off step, the word recognizer either passes the UNK word as is, backs off to a neutral word or backs off to a more general word recognition model trained on a larger, less specific corpus.
In the work, \cite{keller2021bert} authors demonstrated the limitations of spell checker for perturbation identification \& correction. They proposed a method in which context independent probability distribution are created by segmenting the perturbed sentence using BERT tokens and modified version of levenshtein distance. For context dependent probability - all the embeddings of context-independent hypothesis are clubbed into a weighted embedding. Now a token is masked and MLM is used to predict the tokens. This process is repeatedly done for the best approximation. Now, these hypotheses are sent to GPT for getting the language modeling score and the best hypothesis is selected from that. They compared their method again Pyspellchecker, human annotations and RNN trained for spell checking. In the work, \cite{fan2021defending}, authors presented backdoor attacks as adversarial attacks during training of the model and proposed attacking methods for NLG model by inserting trigger words in the input sentence. They further proposed defense strategies by detecting of hacked inputs and output correct results and  preserving the correct input and giving its output. 

In \cite{zhou2019learning} proposed a novel method for perturbation identification and correction, in which they try to recover the perturbed token based on the context and with the help of small world graphs. First they use, BERT model to get the contextualised embedding vector for each token and then pass it to a binary classifier for classification of perturbation. Later they used a BERT-based context network, to be used as the context for predicting the perturbed word. The perturbed word is masked and passed to the BERT to get the embedding of the mask token. Using the embedding vectors and small world graphs they recovered the affected tokens. In the direction of perturbation identification and correction, In \cite{bhalerao2022data} various types of perturbed text are identified and corrected using rule based methods such as alternating characters defense which corrects the combined unicode, space separation, and zero-width space separation perturbation in the entire document. Another rule-based defense used is Unicode Canonicalization which corrects \& replaces unicode and tandem character obfuscation perturbations. They further used a continuous bag of words based embeddings and identified embeddings which are generated by parts of a single word combining random spaced words followed by a process to find similar embedding for words with similar spelling. A skip-gram model is trained with vectors of similar context to ensure the embeddings having similar spelling and context are closer. They evaluated their embeddings with the downstream task of classifying the Facebook posts as engagement bait or otherwise.  In a different line, the work \cite{zhu2021treated} proposed a universal perturbation detection method, TREATED to defend against various perturbation levels without making any prior assumptions. They utilized several reference models to make different predictions about clean and adversarial examples and block them if found adversarial. They designed the reference models on the basis of their consistency on the clean and adversarial data. 
In the direction of identifying perturbations for other language text than English, the work \cite{li2020textshield} proposed a defense model for text classification for Chinese language. Adversarial perturbations are detected in $3$ steps. Neural Machine Translation (NMT) model is used for removing the noise in the input text. The corrected text is converted into multimodal embeddings (semantics, glyph, and phonetics) and the extracted features are given into text classification.

Table \ref{tab:perturubation_id} demonstrates the summary of the various perturbation identification and correction methods. It shows the type of attack for perturbing the input data, the strategy used for detecting the perturbations, NLP applications on which the proposed method is demonstrated, and whether or not they are attempting to correct the perturbed input after their detection. As it can be seen that a large part of perturbation detection and correction method is limited to synonym substitution and misspelling-based adversarial attacks. Also, the demonstration of the proposed defense is largely demonstrated on various types of text classification tasks including sentiment classification, news category classification, and topic classification. The commonly used techniques for perturbation correction after their detection includes blocking the perturbed data, generating or predicting the clean text, and replacing it with similar correct words. There are methods proposed in literature which defend the machine learning model by controlling the direction of perturbations in their embedding space. Some of these methods are discussed in the next section.

\begin{table}[]
    \centering
    \begin{tabular}{||c|p{18mm}|p{32mm}|p{24mm}|C{20mm}|C{18mm}||}
    \hline
       \textbf{ Work} & \textbf{Attack} & \textbf{Method }& \textbf{Application} & \textbf{Perturbation Identification} & \textbf{Perturbation Correction } \\
        \hline
     \cite{wangnatural}    & Synonym Substitution  & Maps synonyms to unique encoding & Sentiment, Topic classification& \centering \checkmark & \textemdash \\
     \hline
     \cite{yang2022robust} & Synonym Substitution& Cluster similar embeddings& Sentiment, Topic classification& \centering \checkmark &  \textemdash \\
     \hline
     \cite{zhou2020defense}  & Substitution based& Mixing embeddings of words \& synonyms & Sentiment, , News category  classification& \centering \checkmark& \textemdash \\
     \hline
     \cite{bao2021defending} & Substitution based & frequency aware randomization & Sentiment classification, Natural Language Inference & \centering \checkmark&  \textemdash \\
     \hline
     \cite{wang2021randomized} & Synonym substitution & Randomized synonym substitution \& vote& Sentiment, Topic, News category classification & \centering \checkmark&  \textemdash \\ 
     \hline 
      \cite{swenor2022using} & Synonym substitution & Random perturbation defense & Sentiment analysis & \centering \checkmark& \textemdash \\
      \hline
      \cite{xie2022identifying} & Attacks from TextAttack \cite{morris2020textattack,zeng2020openattack} & TCAB attack identification dataset \& Text, Language, Classifier properties& Sentiment, Abuse/No-abuse classification& \checkmark& ---\\
      \hline 
      \cite{wang2021textfirewall} & Word level perturbations & Finding word importance & sentiment classification&\centering \checkmark& \textemdash \\
      \hline
      \cite{yoo2022detection} & Word level perturbations & Feature density estimation& Sentiment, News categories, Topic classification, &\centering \checkmark & \textemdash \\
      \hline
      \cite{sakaguchi2017robsut} & Misspellings& ScRNN- Spell checker & ---&\centering \checkmark& \checkmark \\
      \hline 
      \cite{pruthi2019combating} & Misspellings & ScRNN with Backoff& Sentiment Analysis &\centering \checkmark& \checkmark \\
      \hline 
       \cite{keller2021bert}  & Misspellings, orthographic attacks & Context independent probability distribution& Restoring sentences&\centering \checkmark & \checkmark \\
      \hline
    \cite{fan2021defending}  & Backdoor attacks  &Trigger word manipulation and BERTScore&  Machine translation, Dialogue Generation &\centering \checkmark& \checkmark \\
      \hline 
      \cite{zhou2019learning} & Word perturbations& Recover perturbed tokens with small world graphs& Text classification&\centering \checkmark & \checkmark \\
      \hline
      \cite{bhalerao2022data} &  Misspellings & Embeddings similar to original words, Rule based methods& Engagement Bait Classifier&\centering \checkmark &\checkmark \\
      
      \hline
      \cite{zhu2021treated}  & Synonym substitution, replacement order strategy & Using reference models& Sentiment analysis &\centering \checkmark&  \checkmark \\
      
      \hline 
     \cite{li2020textshield} & Word level perturbations & NMT model is used for removing noise & Text classification& \centering \checkmark& \checkmark \\
      \hline 

    \end{tabular}
    \caption{ Summary of defense schemes proposed for perturbation identification and correction}
    \label{tab:perturubation_id}
\end{table}

\subsection{Perturbation direction control}
The proposed work under perturbation direction control alters the direction of the perturbations towards the cleaner text input limiting the adversarial space. 
Along this line, the work \cite{sato2018interpretable} proposed an interpretable adversarial training method by restricting the direction of adversarial samples. The direction of perturbation is restricted to the words in the existing vocabulary so that perturbations could be interpreted even after adversarial training. In the work \cite{zhang2020deep} authors propose to use CBOW to predict the perturbation direction while trying to preserve the meaning of the sentence by placing a constraint on the perturbation direction.
Another work, \cite{rosenberg2021sequence} proposed an adversarial defense mechanism, Sequence Squeezing, aimed to make RNN models and their variants robust against adversarial attacks. The proposed method generates semantic preserving embeddings which are low in the number of features than the original embedding. The squeezed embedding is tested for adversarial attacks in malware detection and added to the training data while diminishing the adversarial space for generating perturbations. Table \ref{tab:pertdir} presents the summary of perturbation direction control based adversarial defense methods in NLP, where each row shows the NLP applications and the type of perturbation used in the threat model in the associate work. 
Proposing adversarial defense for text input data with perturbation direction control is a step towards developing more interpretable defense model than conventional methods of adversarial training. The discussed methods in this category demonstrate  their defense scheme on various type of text classification tasks, such as sentiment classification, malware classification.  Another direction of adversarial defense methods in NLP propose to provide a certified region of robustness while training their machine learning model. Some of these methods are discussed in detail in the coming section. 


\begin{table}[]
    \centering
    \begin{tabular}{||c|c|c||}
    \hline
       \textbf{Work}  & \textbf{NLP Taks} & \textbf{Perturbation} \\
       \hline
       \cite{sato2018interpretable} & Sentiment classification & Word embeddings \\
       \hline
        \cite{zhang2020deep} & Sentiment classification & Word embeddings \\
        \hline
        \cite{rosenberg2021sequence} & Malware detection & Word embeddings for API call command \\
        \hline
        
    \end{tabular}
    \caption{Summary of perturbation direction control based adversarial defense methods in NLP}
    \label{tab:pertdir}
\end{table}


\section{Robustness by certification}
\label{sec:certification}
The methods discussed in the previous sections for adversarial defenses involved word/character substitution-based adversaries where words are synonyms to make the perturbation look indistinguishable. Other methods tweak words by inserting characters, changing spellings, and deleting/swapping characters. All these adversarial samples are necessary for defending the models but they are not sufficient. An attacker can generate millions of adversarial examples by modifying every word in a sentence. A defense algorithm based on adversarial training requires a sufficient amount of adversarial data to increase the robustness, which still does not cover a lot of unseen cases which are generated by exponential combinations of different words in a text input. Also, perturbation control-based methods require identification of perturbations on the basis of already-seen perturbations with a prior assumption of the type of attack. These methods have limitations in their performance when model is exposed to new adversarial instances. Hence, there is a separate set of adversarial defense methods in the literature which are driven by ``certification". These methods train the model to provide an upper bound on the worst-case loss of perturbations and hence provide a certificate of robustness without exploring the adversarial space.

\subsection{Interval Bound Propagation based methods}
Interval Bound Propagation (IBP) \cite{gowal2018effectiveness} is a bounding technique, extensively used in images for training large, robust, and verifiable neural networks. Training the neural networks with IBP technique tries to minimize the upper bound on the maximum difference between the classification boundary and input perturbation region. IBP lets you include the loss term in the training, using which the last layer of the perturbation region can be minimized and kept on one side of the classification boundary. Now, this adversarial region is tighter enough and can be said certified robust.

In this line, the work \cite{jia2019certified} proposed certified robust models while providing maximum perturbations in text classification. Authors used interval bound propagation to optimize the upper bound over perturbations. IBP gives an upper bound over the discrete set of perturbations over word vector space. IBP computes an upper bound on the model’s loss when given an adversarially perturbed input. This bound is computed in a modular fashion.
In another work \cite{huang2019achieving} introduced a verification and verifiable training of neural networks in NLP. Authors proposed a tighter over-approximation in the form of a ‘simplex’ in embedding space in the input to generate perturbations. To make the network verifiable they define it as the convex hull of the all the original unperturbed inputs as a space of delta perturbation. Using IBP algorithm they generated robustness bounds (by generating bounds for each layer). 
In the work, \cite{ye2020safer} proposed structure-free certified robust models which can be applied to any arbitrary model. This method overcomes the limitations of IBP based method in which they are not applicable to character level and sub-word level models. They prepared a perturbation set of words using synonym sets, top-K nearest neighbors under the cosine similarity of GLOVE vectors, where K is a hyperparameter that controls the size of the perturbation set. They further generated sentence perturbations using word perturbations and trained a classifier with robust certification. In the context of IBP methods, \cite{wallace2022does} demonstrates the lack of generalizability of IBP-based methods for novel contextual embeddings and a wider range of NLP tasks. They demonstrated the performance of their method in the sentiment analysis task.


\subsection{Certified robustness for RNN networks}
Despite having a plethora of work in finding a certificate for robustness, there is a lack of applicability in RNN based network due to their inherent complexity. Hence, in another line of work for robustness by certification, certified robustness for RNN based networks and self-attentive networks is proposed. 
In this line, the work Popqorn \cite{ko2019popqorn} proposed certified robustness for RNN based networks such as LSTM, GRUs. The challenge is to find a certificate of robustness in RNN based networks with their complex feedback architectures, the sequential inputs, and the cross-nonlinearity of the hidden states. Authors used 2D planes to bound the cross nonlinearity in LSTMs and proposed to find a certificate within a $l_p$ ball (attack distance) if the lower bound on the true label output unit is larger than upper bounds of all other output units. They generated certificate of robustness  by writing all the bounds as a function of epsilon and tried to find the optimum value of epsilon using a binary search procedure. 
In the work Cert-RNN \cite{du2021cert}, they overcame the limitations of Popqorn \cite{ko2019popqorn} by a robust certification framework for RNNs. The method overcame the limitations of Popqorn by keeping the inter-variable correlation and speeding up the non-linearities of RNN for practical uses. Authors created a zonotope \cite{eppstein1995zonohedra} around the input perturbations and used that to be passed through a vanilla RNN or LSTM. The properties of the output zonotope can be verified to be certifiably robust. They used zonotope instead of a box to preserve inter-variable correlation, the precision of the network, and achieve a tighter bound. They could achieve tighter bounds and at least 19 times faster framework than Popqorn.
In this line, \cite{zhang2021certified} proposed a novel approach, Abstractive Recursive Certification (ARC) for certified robustness in RNN based networks. Authors defined a set of programmatically perturbed string transformations and constructed a perturbation space using those transformations proposed in \cite{zhang2020generating}. They memorized the hidden states of the strings in the perturbation space that shared a common prefix to reduce the evaluation of LSTM cells while finding an upper bound to the loss and avoiding re-computing of hidden states. Following that they represent all the perturbation sets as a hyperrectangle and pass the hyperrectangle through the remaining network using IBP technique \cite{gowal2018effectiveness}. Following a similar direction, the work in \cite{ryou2021scalable} presents Polyhedral Robustness Verifier of RNNs (PROVER) which represents the perturbations in input data in the form of polyhedral which is passed through a LSTM network to obtain a certifiable verified network for a more general sequential data.
Another work in this direction is proposed by \cite{bonaert2021fast} where authors proposed DeepT an abstract transformer-based network certification method. They attempted to certify larger transformers against synonym replacement-based attacks. In this work, authors propose to use multi-norm zenotopes improving the precision of standard zenotope based methods which works well for longer sentences by certifying larger radii of robustness ($\times28$ of existing methods).
In another work \cite{shi2020robustness} proposed an algorithm for verifying the robustness of transformers with self-attention layers which include challenges such as cross-linearity and cross-positional dependency. They provide a lower bound to a boundary (delta certificate) which will be always greater than 0 (probability of correct class is always higher than incorrect class) within a set of inputs which also include perturbations and tighter than IBP. They achieved this by computing lower/upper bound for each neuron with respect to the input space.

\subsection{Other convex optimization based methods}
There are other methods in literature for finding certified robustness for neural networks which used several convex optimization schemes and randomized smoothing-based schemes. 
In this line, \cite{steinhardt2017certified} certified defence method is proposed for text classification task. They consider data sanitation defences, which examine the entire datasets and try to remove poisoning points. They upper bound the worst possible test loss of any attack which works in an attacker-defender setting at the same time. They generated a certificate of robustness (upper bound) by inserting perturbed data at the time of training where defender is learning to remove outliers at each iteration. Upper bound fits all possible points that evade outlier removal. In the work \cite{raghunathan2018certified} authors proposed certified robustness method based on semi-definite relaxation. They computed an upper bound on the worst case loss of the neural networks with one hidden layer. The computed certificate of robustness provides an upper bound on the robustness for all kinds of attacks and being differentiable they trained it jointly with the network. The work \cite{wang2021certified} provided a certificate of robustness with the idea of differential privacy in the input data. They implemented differential privacy in the textual data by treating a sentence as a database and words as an individual records. If a predictive model satisfies a certain threshold (epsilon-DP) for a perturbed input, its input should be the same as the clean data. Hence providing a certification of robustness against L-adversary word substitution attacks. 


 \begin{table}[h]
    \centering
    \begin{tabular}{||p{18 mm}|c|p{40mm}|p{40mm}|p{25mm}||}
    \hline
      \textbf{Certification method}   &   \textbf{Work} & \textbf{NLP Taks} & \textbf{Models} & \textbf{Perturbations}\\
      \hline
        \multirow{4}{*}{IBP}  & \cite{jia2019certified} & NLI, Sentiment classification& Feed Forward network, CNN, Bi-Directional LSTM, Decomposable attention& Word substitution\\ \cline{2-5}
       & \cite{huang2019achieving} & Sentiment \& Topic Classification& 1 layer convolution network & word substitution and character typos \\  \cline{2-5}
       & \cite{wallace2022does} & Sentiment classification & CNN & word embeddings\\ \cline{2-5}
       & \cite{ye2020safer} & Sentiment \& Text classification & BERT & Word Substitution \\  \hline
      \multirow{4}{*}{RNN based} & \cite{ko2019popqorn}& Question Classification & LSTM, GRU & $\epsilon$ bounded $L_p$ ball \\ \cline{2-5}
      & \cite{du2021cert}  &  sentiment analysis, toxic comment detection, and malicious URL detection & RNN, LSTM & $\epsilon$ bounded $L_p$ ball\\ \cline{2-5}
      & \cite{zhang2021certified} & Sentiment Classification & LSTM & Word substitution \\ \cline{2-5}
      & \cite{ryou2021scalable}  & Speech classification & RNN, LSTM & $\epsilon$ perturbation, $dB$ perturbation \\ \cline{2-5}
      & \cite{bonaert2021fast} & Sentiment Classification & Transformer networks & $L_p$ noise where $p \in \{1,2, \infty \}$ \\ \cline{2-5}
      & \cite{shi2020robustness} & Sentiment classification & Transformer & $\epsilon$ bound perturbations \\ \hline
     & \cite{steinhardt2017certified} & Sentiment Classification& SVM & $\epsilon n $ poison points  \\ \cline{2-5}
      & \cite{wang2021certified} & Text classification & LSTM & Word substitution  \\ \cline{2-5}
   \multirow{4}{*}{Other methods }     & \cite{zeng2021certified}  & Sentiment \& Topic classification & BERT \& RoBERTa & Word substitution \& Character level \\ \cline{2-5}
    &  \cite{la2020assessing} & Sentiment, News, Topic classification & CNN, LSTM & Word substitution  \\ \cline{2-5}
      &  \cite{pruksachatkun2021does} & Toxicity, Occupation classification & CNN, BERT & Word substitution  \\ \hline
    \end{tabular}
    \caption{Summary of the certifiable robustness methods in NLP }
    \label{tab:certified}
\end{table}

Another work, \cite{zeng2021certified} proposed defense algorithm to overcome the limitations of previous methods with an assumption that perturbation generation methods will be known a priory. They proposed RanMASK, a certifiably robust defense method against text adversarial attacks based on a new randomized smoothing technique for NLP models. Manually perturbed input text is given to the mask language model. Random masks are generated in the input text in order to generate a large set of masked copies of the text.  A base classifier is then used to classify each of these masked texts, and the final robust classification is made by “majority vote” and trained with BERT and RoBERTa to generate and train with masked inputs. 
Another work in this direction \cite{la2020assessing} estimated the Maximum Safe Radius (MSR) for a given input text, i.e. minimum distance between the classification boundary and embedding space. They quantified the robustness of neural networks against word replacement which is based on a minimum safe radius. They approximated the upper bound using Monte Carlo tree search and the lower bound by constraint relaxation technique of MSR for CNN and LSTM networks. In \cite{pruksachatkun2021does}, the authors also tried to club the concept of fairness and robustness to increase the robustness of a neural network. They demonstrated that a certified robust model can also be used as a bias mitigation system to build trustworthy NLP systems. They integrated a bias mitigation system with state-of-the-art certified robust models to improve the robustness of a model. Table \ref{tab:certified} presents the summary of the robustness by certification methods in NLP where rows are grouped by the certification method used, and each row describes the NLP application, machine learning model used for certification and type of perturbations generated in the threat model with each associated work. Currently these methods are not proven to be generalized across different types of deep neural networks and they have been evaluated for small set of NLP tasks and on smaller networks. 
There are various methods for defending the neural network from adversarial attacks and achieving robustness which are not discussed in these sections and follows a different line of approach, described in the next section.

\section{Miscellaneous}
\label{sec:misc}
In the previous section, various methods are discussed for adversarial defenses and robustness enhancement. These methods follow the proposed taxonomy and categorization of adversarial defence schemes discussed in Sec. \ref{sec:taxonomy}. However, there are several other schemes proposed in recent years that do not fall into any of the categories discussed above.
In the direction of enhancing robustness by bias reduction, the work \cite{stacey2020avoiding} tries to remove hypothesis-only bias for NLI datasets by using adversarial classifiers to detect bias  in the sentence representation. They demonstrated that the larger the sentence embeddings, the harder it is to remove the bias and requires more adversarial classifiers. They tested models with 1 and 20 classifiers where 8 out of 13 datasets performed better with 20 classifiers and for 3 of them 1 and 20 gave the same performance. 

In another line of defending APIs from adversarial attacks the work \cite{he2021model} showed that hosted BERT-based APIs are vulnerable to theft and users can query the API for a dataset and train a BERT model to replicate the API. The replicated model can then be used for adversarial example transfer. They suggested a parameter-based defense strategy by using a temperature parameter in softmax to smooth the output prediction probabilities. They further add perturbation noise with variance sigma to the output probabilities where the larger the variance stronger the defense.

In the direction of creating various adversarial examples for adversarial training, the work \cite{guo2021towards} proposed variable length adversarial attack in contrast to an existing method  which focuses on fixed length. This is achieved by using a special ``BLK" token during fine-tuning and then using 3 atomic operations addition, deletion \& replacement to create adversarial examples. They show that this method successfully attacks the models in NLU and NAT tasks and demonstrated its use for creating augmented data for adversarial training. In the same line, the authors in \cite{fan2021defending} presented backdoor attacks as adversarial attacks during training of the model for NLG models. They proposed post-hoc defense against the attacks by using token removal and token substitution on a sentence and corpus level. 

Taking VAT in different directions, \cite{zuo2021adversarial}  proposed a novel strategy, Stackelberg Adversarial Training (SALT) which employs a Stackelberg game strategy. There’s a leader which optimizes the model and a follower which optimizes the adversary. In this Stackelberg strategy, the leader is advantageous knowing the follower’s strategy and this information is captured in Stackelberg gradient. They find the equilibrium between the leader and follower using an unrolled optimization approach. 

Another work in the direction of robustness enhancement, proposed in \cite{jones2020robust} introduced Robust Encodings (RobEn), which is a simple framework that guarantees robustness, without making any changes to model architecture. The core component of RobEn is an encoding function, which maps sentences to a smaller, discrete space of encodings on a token level. They attempt to cluster all possible adversarial typos into a single cluster using graph-based agglomerative  clustering and try to balance between having too many words in a cluster versus a single word in a cluster. In the same line, for languages other than English, \cite{li2021enhancing} proposed AdvGraph to enhance the adversarial robustness of Chinese NLP models with modified embeddings. Due to the inherent complexity in Chinese language, the existing adversarial defense models are difficult to be extended for the Chinese language, hence they propose to capture the similarity in words using graphs. They constructed undirected adversarial graph based on the glyph and phonetic similarity of Chinese characters and learned the representations through graph embeddings to be used with semantic embeddings to be used for other downstream tasks.

There are various metrics used in the literature for evaluating the robustness of the models against adversarial attacks. Evaluating adversarial robustness is equivalent to evaluating the performance of a model in the presence of adversarial attacks, before and after the implementation of defense mechanism. Hence, some of these metrics are standard performance metrics widely used in machine learning literature. Other variants of evaluation metrics are proposed to measure the performance of certifiably robust models. The next section describes these evaluation metrics used in adversarial defense literature.

\section{Metrics for evaluation}
\label{sec:metrics}
There are various evaluation metrics that are extensively used in literature for evaluating the proposed defense methods. Majorly these metrics are performance evaluation metrics for the machine learning model which is required to be defended. Adversarial defense methods are evaluated with the performance of the model after the defense method is implemented or run-time evaluation in case of methods that aim towards optimizing a lower/upper bound. Hence, some of these metrics are accuracy, error, loss analysis, measurement of the success of adversarial attacks, similarity with ground truth in the case of language generation etc. In this section, some of these commonly used metrics are described in detail.

\begin{itemize}
    \item \textbf{Prediction accuracy (Conventional Accuracy)}: Adversarial defense methods for text classification models, such as sentiment classification, Natural Language Inferencing tasks, are evaluated on the basis of the prediction accuracy after implementation of defense algorithm. The prediction accuracy after defense is compared with prediction accuracy after attack, and if there is a surge in accuracy, the defense method is considered to be successful. It is defined as the fraction of test set that is correctly classified \cite{wang2021certified}. Conventional accuracy is a standard metric that is used to evaluate any deep learning system and it can be used to evaluate any defense method.
    \begin{equation}
        \frac{\sum_{t=1}^T CorrecClass(X_t,L, \epsilon)}{T}  \nonumber
    \end{equation}
    Here, $CorrecClass(X_t,L, \epsilon)$ gives 1, if test sample, $X_t$ is correctly classified for test data $T$. 
    
    \item \textbf{Loss function analysis}: The negative log-likelihood (loss function) is tested over its rate for adversarial training as regularization, and virtual adversarial training-based methods. It can indicate lower error rate and reduced overfitting in adversarial training based regularization.  
    
    \item \textbf{Error analysis}: Adversarial defense methods are also evaluated on the error rate in the prediction of the model. The error rate is compared with an adversarial attack before and after the implementation of defense schemes. A lesser error rate after defense method, entails its successful defense scheme.  Error analysis is a standard metric that is used to evaluate any deep learning system for prediction and it can be used to evaluate any defense method in literature.  
    
    \item \textbf{Embedding testing}: Embedding test is done for evaluating the embeddings generated for adversarial training. Similarity metrics such as Edit distance, Jaccard similarity coefficient, and semantic similarity metrics are used to evaluate the utility of the adversarial samples generated by finding their similarity with the original input samples. 
    
    \item \textbf{Human Evaluation}: To measure the utility of the adversarial samples for adversarial training, human evaluation is also performed in literature \cite{garg2020bae}. Human annotators are asked to judge the adversarial examples in terms of their naturalness by presenting both original and adversarial examples. The good quality adversarial examples are used in adversarial training and further robustness of the deep neural network is evaluated. Adversarial attacks and defense methods are closely associated, hence weaker the adversarial attack, the stronger the defense strategy. 
    
    \item \textbf{Attack Success Rate (ASR)}: ASR \cite{wu2021performance} is a measure of success of the adversarial samples created for the potential adversarial attack. This metric is used to measure the effectiveness of the adversarial attack after any of the defense scheme is implemented. The attack success rate is measured before and after defending the model and a drop in its value will imply a more robust model.  It is defined as:
    
    \begin{equation}
        ASR= \frac{N_{successful}}{N_{Total}} *100  \nonumber
    \end{equation}
    Where, $N_{successful}$ is the number of adversarial samples that were able to successfully fail the model, and  $N_{Total}$ is the total number of adversarial samples generated. 
    
    \item \textbf{BLEU}: Adversarial defense schemes for Natural language generation models utilize performance metrics such as BLEU score \cite{papineni2002bleu} for the evaluation of their proposed method. BLEU score is measured using $n$-gram to evaluate the quality of the generated natural language by comparing it with ground truth. It is defined as:
    \begin{equation}
 p_n=\frac{\sum_{C\in \{Cand\}}{\sum_{gram-n\in C}{Count_{clip}(gram-n)}}}{\sum_{C'\in\{Cand\}}{\sum_{gram-n'\in C'}{Count(gram-n')}}} \nonumber
\end{equation}
 \begin{equation}
    BP= 
    \begin{cases}
      1, & \text{if}\ c>r \\
      e^{\frac{1-r}{c}}, & c\leq r   \nonumber
    \end{cases}
  \end{equation}
    
    \begin{equation}
    BLEU=BP.exp^{\sum_{i=1}^{N}{W_n}log(p_n) } \nonumber
\end{equation}

  Where $(p_n)$ is the n-gram modified precision score, BP is brevity penalty used for longer candidate summaries and for spurious words in it, $c$ is the length of the candidate summary, and $r$ is the length of the reference summary.
  
    \item \textbf{Number of Queries}: This denotes average number of times attacker queries the model. The higher the average number of queries made by an attacker, more difficult is to fail a defense mechanism \cite{li2021searching}. It can be used for evaluating any adversarially trained defense model using adversarial instances or by controlling the perturbations.   
  
    \item \textbf{Precision on certified examples}: Precision on certified examples \cite{lecuyer2019certified}, which measures the number of correct predictions exclusively on examples that are certified robust for a given prediction robustness threshold. It is defined as:
 
    \begin{equation}
        Precision= \frac{\sum_{t=1}^T (isCorrect(X_t)\;\& \;robustSize(p_t, \epsilon, \delta, L) \geq Threshold}{\sum_{t=1}^T robustSize(p_t, \epsilon, \delta, L)\geq Threshold}        \nonumber
    \end{equation}
    
    Where, $isCorrect(X_t)$ gives a value of $1$ if the input sample $X_t$ is correctly classified, and $ robustSize$ gives the certified robustness value for the bound $L$. 

    \item \textbf{F1 Score}: This is a metric that combines precision and recalls both, to evaluate the overall performance of the model. It is used for comparison of model's performance before and after defense mechanism implementation. F1 score is a standard metric used for evaluating the performance of deep neural networks and it can be used to evaluate any defense method. 
    It is defined as:
    \begin{equation}
        F1\;Score= \frac{2*(Precision * Recall )}{(Precision + Recall )}    \nonumber
    \end{equation}
 
    \item \textbf{Certified Radius}: In certification based adversarial defense methods, robust radius is the largest radius centered around input sample $X_t$, for which the classifier does not change its value for its corresponding perturbed sample $X_t^{adv}$.
    However, calculating robust radius of a deep neural network is a NP-hard problem \cite{zhai2020macer}. Hence, certification based methods are tested for their certified radius for different norms of perturbations for targeted model on parameters such as minimum radius, average radius, and time taken to obtain it. Certified radius is a lower bound to the robust radius and leads to a guaranteed upper bound of the robust classification error. 

    \item \textbf{Certificate Ratio (CR)}: This metric is used in certification based defense schemes. It is the fraction of testing samples, that satisfies the certification criteria after prediction \cite{wang2021certified}. It is defined as:
    
    \begin{equation}
        CR= \frac{\sum_{t=1}^T CertifiedCheck(X_t,L,\epsilon)}{T}    \nonumber
    \end{equation}
    Here, $CertifiedCheck(X_t,L,\epsilon)$ gives $1$, if the fraction of the test data is certified robust. 
    
    \item \textbf{Certified Robustness}: This metric is used in certification-based defense schemes. Certified Robustness \cite{levine2020robustness} for a particular $X_t$ is the maximum value $\rho$ for which it is certified that classifier will return the correct label where $X_t^{adv}$ is its corresponding perturbed sample, such that $||X_t- X_t^{adv}|| \leq \rho$. 

    \item \textbf{Median Certified Robustness}: This metric is also used in certification-based defense schemes. The Median Certified Robustness \cite{levine2020robustness, zeng2021certified} on a dataset is the median value of the certified robustness across the dataset. It is the maximum value $\rho$ for which the classifier can guarantee robustness for at least 50\% samples in the dataset. In other words, we can certify the classifications of over 50\% samples to be robust to any perturbation within $\rho$.

    \item \textbf{Certified accuracy}: This is the metric used for evaluating the certifiable robust models. Certified accuracy \cite{lecuyer2019certified, wang2021certified} is the percentage of correct test samples for a certified robust model for the given perturbation. It denotes the fraction of testing set, on which a certified model’s predictions are both correct and certified robust for a given prediction robustness threshold. It is defined as:    
    \begin{equation}
        Certified\;Accuracy= \frac{\sum_{t=1}^T (isCorrect(X_t)\; \& \; robustSize(scores, \epsilon, \delta, L) \geq Threshold}{T}    \nonumber
    \end{equation}
    Where $robustSize(scores, \epsilon, \delta, L)$ is the certified robustness size for the bound $L$ and $isCorrect(X_t)$ give a value of $1$ if the input sample $X_t$ is correctly classified for test data $T$. 

    \item \textbf{Conditional Accuracy}: This is the metric used for evaluating the certifiable robust models. Conditional Accuracy is proposed by \cite{wang2021certified}, evaluating the classification accuracy of both a clean sample $X_t$ and its corresponding adversarial sample $X_t^{adv}$ withing a bound $L$. It checks when $X_t$ is certified within bound L, whether $X_t^{adv}$ is also classifying correctly. It is defined as:
    
    \begin{equation}
        Conditional\;Accuracy= \frac{\sum_{t=1}^T (Certified(X_t,L, \epsilon)\; \& \; corrClass(X_t^{adv},L, \epsilon) }{\sum_{t=1}^T (Certified(X_t,L, \epsilon)}    \nonumber
    \end{equation}
    Where, $Certified(X_t,L, \epsilon)$ gives $1$, when clean input sample $X_t$ is successfully certified and $corrClass(X_t^{adv},L, \epsilon)$ gives $1$ when its perturbed input $X_t^{adv}$ is also correctly classified.

    \item \textbf{CLEVER score}: Cross Lipschitz Extreme Value for nEtwork Robustness (CLEVER) score is proposed by \cite{weng2018evaluating} is a novel robustness evaluation metric. It is attack-independent and can be applied to any arbitrary neural network classifier and scales to large networks. CLEVER metric is an estimation of local Lipschitz constant which represents ``lower bound of the robustness in input data" or minimum amount of perturbation required to a natural sample to fail a classifier. The increased clever score indicates that the network is indeed made more resilient to adversarial perturbations after any defense mechanism is used.
\end{itemize}

\section{Adversarial datasets and frameworks }
\label{sec:datasets}
There are several dataset in NLP which are proposed for adversarial evaluation. One such dataset is DailyDialog++ \cite{sai2020improving}, which is an extension of DailyDialog dataset \cite{li2017dailydialog} and adversarial dialogue generation dataset. DailyDialog++ contains, 5 additional relevant and adversarially irrelevant responses, for 11k context conversation derived from DailyDialog. This dataset is used for evaluating robustness of dialogue generation models with adversarial examples in the dataset against adversarial attacks. Further these adversarial examples are used in adversarial training to imporve the performance of dialogue generation models. 
Another work is ANLI \cite{nie2019adversarial}, which proposed adversarial dataset for natural language inference systems. ANLI composed of adversarial examples for NLI collected in 3 iterative rounds having human and machine in loop.  ANLI consist of  $103k$ examples of sentences,  starting with short multi-sentence passages from Wikipedia and having annotators writing adversarial hypothesis. In this process, they tested these samples with state-of-the-art NLI models and got then verified by human annotators hence proposing human-and-model-in-the-loop enabled training (HAMLET) scheme for data collection. The adversarial examples collected in this dataset are used in adversarial training to improve the performance of Natural Language Inferencing models. 

In the same line, authors in \cite{li2021adversarial} proposed a novel large scale dataset adversarial VQA for visual question answering task using the HAMLET scheme proposed in \cite{nie2019adversarial}. In this work they presented an image to an annotator and ask them to write a tricky question that could fool a model. Hence they iteratively collected $243.0K$ questions for $37.9K$ images by having humans and models competing in the loop in $3$ rounds. This dataset is used in evaluating the robustness of SOTA VQA models. Authors have further used this dataset for data augmenting, for the purpose of adversarial training and demonstrated a higher performance on robust VQA benchmarks.
For the purpose of evaluating adversarial examples for question answering task, authors in \cite{jia2017adversarial} proposed, Adversarial SQuAD dataset, that contained adversarially inserted sentences. These sentences are automatically generated in a concatenative manner, without changing the meaning of the paragraph or question. Fake answers to these questions are also generated with same POS type. This dataset is used for evaluating the performance of various state-of-the-art question answering models. The best performing model with adversarial examples is analysed for the features causing the high performance. Based on this analysis, additional features were incorporated in the model, separately and in combination, to enhance its performance both with and without adversarial examples. Authors further used this dataset for adversarial training of modified model to validate its stability. In the same line, \cite{wallace2019trick} proposed a question-answering test bed Quizbowl, using a Human-In-the-Loop framework. In this work, human authors are asked to write adversarial questions which are designed to fail state-of-the-art question answering models appearing ordinary to human. This test dataset is used for adversarial evaluation of SOTA question answering models. 

In another work, \cite{zhang2019paws} adversarial dataset, namely, Paraphrase Adversaries from Word Scrambling (PAWS), for paraphrase detection is proposed. PAWS is generated from sentence in Quora and Wikipedia, where adversarial samples are generated using language model based controlled word swapping and back translations. PAWS dataset is used for adversarial evaluation of paraphrase detection models and measured sensitivity of these models on word order and syntactic structure. Authors further used this dataset for adversarial training of state-of-the-art paraphrase detection models.
In the direction of perturbation identification \cite{xie2022identifying} proposed a dataset Text Classification and Attack Benchmark (TCAB), which was created for the purpose of detecting and labeling the textual perturbation in the input text. TCAB is an extensive dataset containing $1.5$ million attack instances, generated using $12$ attacks from the toolkits \cite{morris2020textattack, zeng2020openattack} which are targeting $3$ classifiers, trained on $6$ domain datasets of sentiment and abuse classification. There is a total number of $216$ attacks taken into consideration in TCAB dataset which are a combination of different type of attacks. This dataset consists of a total of $1,53,9881$ adversarial examples along with the clean instances taken from the original datasets. They further use text, language, classifier properties to detect the perturbation and type of the attack in the input text. 
Table \ref{tab:datasets} shows the summary of the publicly available adversarial datasets for various NLP tasks. It describes the task for which the dataset is proposed, its statistics and methods used in data collection and annotation.

\begin{table}[h!]
    \centering
    \begin{tabular}{||c|c|c|c||}
    \hline
    \textbf{Dataset} & \textbf{Task} & \textbf{Statistics} & \textbf{Method} \\
    \hline
      DailyDialog ++ \cite{sai2020improving}   & Dialogue Generation & $11k$ context, $5$ responses& Human Annotators\\
      \hline
      ANLI \cite{nie2019adversarial} & NLI systems & $103k$ sentences& Human \& machine in loop \\
      \hline
      Adversarial VQA \cite{li2021adversarial} & Visual Question Answering & $243.0k$ Q \& $37.9k$ images& Human \& machine in loop\\
      \hline
      Adversarial Squad \cite{jia2017adversarial} & Question Answering & $ 107,785$ Q & Machine generated \\
      \hline
      Quizbowl \cite{wallace2019trick} & Question Answering  & $1213$ Q-A pairs & Human \& machine in loop \\
      \hline
      PAWS \cite{zhang2019paws} & Paraphrase detection & $ 108,463$ paraphrase pairs &  Machine generated\\
      \hline
      TCAB \cite{xie2022identifying} & Sentiment, Abuse classification & $1.5m$   Adversarial ins. &  Machine generated \\ 
      \hline
    \end{tabular}
    \caption{Summary of adversarial datasets in NLP, here NLI = Natural Language Inference, Q= Questions, A= Answers, ins= instances}
    \label{tab:datasets}
\end{table}


There are papers in the literature that proposed python frameworks for a complete adversarial evaluation for several NLP tasks with various attack algorithms. One such work is TextAttack \cite{morris2020textattack} which is a python framework for end-to-end adversarial evaluation with 16 different adversarial attack methods. It consists of a task-specific goal function, data augmentation schemes along with perturbation constraints that validate the perturbation with original inputs, and a repetitive model querying search system. It facilitates the user to benchmark existing attacks, and create novel attack schemes by using new and existing components and evaluating them. Improving the shortcomings of TextAttack framework, Open attack \cite{zeng2020openattack} is proposed, which included $15$ different types of attacks such as sentence, word, character level. It also supports Chinease in addition to English language models and supports  multi-process running of attack models to improve attack efficiency. In the same line another evaluation framework \cite{xu2020elephant} ``Elephant in the room" is proposed, which consists of a combination of automatic evaluation metrics and human judgments. Targeting the sentiment classification task, it included crowd-sourced human judgments, for judging the naturalness, preservation of original label, and comparing similarity on a text similarity metric. 

\section{ Recommendations for future work}
\label{sec:futurework}
In this paper, an exhaustive survey of the methods proposed in the literature to defend neural networks from adversarial attacks and to enhance their robustness is presented. We proposed a novel taxonomy for adversarial defense mechanisms for various tasks in NLP. Methods for adversarial defense in NLP are broadly divided into three categories, (i) methods based on adversarial training, (ii) methods based on perturbation detection (iii) methods providing a certificate for robustness. Another part of methods which do not follow any of the above mentioned schemes is categorized as miscellaneous. While there is an ample amount of work proposed in this direction for tasks in NLP, there are still various gaps remaining which should be looked at as potential future directions in this area. 

\begin{itemize}
    \item \textbf{Larger part of work based on adversarial training}: A large portion of work in adversarial defenses for NLP revolves around adversarial training and data augmentation. Despite having a plethora of work in this direction, there is a large part of adversarial defense methods which are oriented towards augmenting the data for generating adversarial examples for adversarial training. Adversarial training is undoubtedly a successful defense scheme for adversarial attacks but it lacks generality for a more practical purpose. It makes the model highly robust for a certain kind of attack but it still makes it vulnerable to the type of examples the model has not seen. Hence, the other methods for adversarial defenses should also be explored. 
    
     \item \textbf{Hand-crafted generation of adversaries}:
     As a future work recommendation, more attention should be given to the automatic generation of adversarial examples. Most of the methods based on adversarial training with data augmentation rely on hand-crafted adversarial examples. There are methods that replace words with their synonyms, or adjacent words, flip the character or concatenate words at the end of sentences. Despite being highly efficient, these examples are devised by humans rather than having an automatic generation of adversarial examples and have their own limitations. Hence, more efforts can be put in the direction of the automatic generation of adversarial examples. 
     
     \item \textbf{Interpretability of perturbations in text data}:
     Adversarial examples should be less human perceptible/natural looking to make the model robust in a practical attack scenario. In contrast with adversarial defenses for computer vision based methods and examples generated on images, examples on text for NLP tasks are difficult to generate because of their discrete nature. Modifying pixels in images for generating adversarial examples are less perceptible to human than modifying a character or word in an input text. Hence to make to defense method more useful in a practical scenario, more efforts should be put in this direction.
     
    \item \textbf{Exact robust certificate calculation}:
    In existing literature only the upper and lower bound on the certificate could be calculated, rather than the exact robustness certificate. While adversarial training based defense methods provided a sufficient exposure towards achieving robust neural networks, progressively novel adversarial attacks kept rolling in. A new set of methods to achieve robustness were proposed in the direction of providing a certificate of robustness for a neural network, attempting to put an end to this race. Certification based adversarial defense methods, definitely provides a more generalization for the neural network for a task but certification does not make a sufficient property of a model for achieving robustness. Finding an exact robustness certificate to the set of input is a non-convex optimization problem and is inefficient to solve. However, in literature authors have relaxed this problem to convex optimization by finding an upper or lower bound to the robustness certificate. Despite the efforts towards finding a certified robustness convex optimization can lead to lossy results and there is a scope of finding a tighter bound. Hence future efforts in this direction can be made to improve the tightness of existing robustness certificates. 
    
    \item \textbf{Scalability of certification based robustness}:
    Certification methods, do not scale to large and practical networks used in solving modern machine learning problems. The current certification based robustness method in literature, is implemented on theoretical models on a small scale. They are not scalable to larger and deeper networks for practical purposes. Hence, in future attempts should be made in this direction. 
   
    \item \textbf{Generalization of adversarial training}:
    The current state of the art methods based on adversarial training in NLP are designed in a task specific manner. There is a lack of generality in adversarial example generation schemes which could be used for multiple NLP tasks effectively. Therefore, steps can be taken in this direction in the future.      
    
    \item \textbf{More explainable models in case of inserting perturbations in the loss function}:
    There is a lack of explainability and transparency in the regularization based defense methods. The loss function contains the term responsible for introducing perturbations at the training time. However, there is no explanation for these methods for their correctness or high accuracy. 

    \item \textbf{Interpretable adversarial training}:
    While adversarial training based method covers a larger portion of adversarial defense literature in NLP, these models are hardly interpretable in terms of adversarial instances. Adversarial examples are provably helping these models become more robust toward adversarial attacks, but there are questions such as, ``how they are improving the performance", ``is this list of adversarial instances exhaustive or could there be more such instances", ``is the model robust towards some specific type of attacks or they are able to defend from any type of attacks", are still need to be answered.  
    
    \item \textbf{Attack agnostic perturbation detection}: A better part of the literature in perturbation detection methods works on the prior assumption of spelling-based attacks or synonym substitution based attacks. However, adversarial perturbation in textual data could be caused by multiple types of attacks, individually or in combination. Hence, there is a requirement for perturbation detection schemes that attack agnostic and present a general framework for perturbation detection. 

    \item \textbf{Novel methods to identify the existing perturbations in the input}:
    A large part of perturbation detection schemes depends upon spell checking methods and methods which enumerate or cluster synonyms. In a practical scenario, it is highly inefficient to compute and enumerate synonyms of words for perturbation recognition. Hence, there is a requirement for novel and innovative methods for identifying the perturbations in the input text which do not involve traditional methods such as spell checking, synonym mapping, or their other variants. 
    
    \item \textbf{Better coverage for NLP applications}: 
    The proposed adversarial defense methods evaluate their method on certain NLP tasks to validate strength of their method. In addition to being limited to certain type adversarial attacks, a large part of defense methods demonstrate their algorithm on different types of text classification tasks. These tasks include sentiment, hate, news, topic, abuse, malware classification type of problem which comes under the umbrella usecase of text classification. Text classification is one of the Natural Language Understanding (NLU) task, while there are other NLU tasks such as named entity recognition, machine translation, automatic reasoning etc. having lesser coverage. Moreover, a very limited number of defense schemes have demonstrated their methods on Natural Language Generation (NLG) use cases such as summary generation, question answering. Hence, there is a requirement for adversarial defense methods covering the other NLP applications. 
    
    \item \textbf{Better evaluation metrics}: The current evaluation of robustness against adversarial attacks for NLP models is based on the performance metrics of the actual model, i.e. accuracy, precision-recall, error-analysis, etc. Hence, there is a requirement for novel evaluation metrics that could measure the robustness and ability to defend against adversarial attacks. There is also a requirement for sensitivity metrics for machine learning models, for measuring their sensitivity towards adversarial examples. There also ain't enough ways to evaluate defense mechanisms themselves along the lines of perceptibility and naturalness. Hence, in future, more evaluation metrics can be brought along this line. 

\end{itemize}

\section{Conclusion}
\label{sec:conclusion}
In this paper, a survey is presented for adversarial defense methods for various tasks in NLP. We proposed a novel taxonomy for adversarial defenses in NLP covering a wide range of recently proposed papers. This survey tries to fulfill the gaps in existing surveys where adversarial attack schemes were more focused. However, in recent years, numerous methods are proposed for defending neural networks with adversarial attacks and enhancing their robustness. Coming up with novel defense schemes for advanced NLP systems is as important as coming up with novel attacks to make these neural networks robust and safe for practical purposes. This survey also covers various adversarial datasets, frameworks proposed in recent times for efficient adversarial evaluation of the SOTA models, and evaluation metrics to quantify their robustness. Moreover, it highlights various recommendations for future work considering the limitations and gaps in the existing literature on adversarial defenses. This survey, therefore, provides a strong basis and motivation for future research in developing robust and safe neural networks in NLP tasks. 

\appendix
\section{Various Natural Language Processing tasks}
\label{sec:NLPtasks}
In this section various NLP tasks are described in detail, on which adversarial defense schemes are demonstrated in literature. These tasks include applications of both Natural Language Understanding (NLU) and Natural Language Generation (NLG). 

\subsection{Text classification}
Text classification is an essential NLP task that involves classifying text data into multiple categories. It encompasses a vast array of problems, including but not limited to sentiment analysis, natural language inference, malware detection, spam filtering, and topic labeling. In the text classification process, a contextual representation of the input text is generated, followed by the selection of an appropriate classifier \cite{aggarwal2012survey,li2020survey,kowsari2019text}. This process may also involve preprocessing and intermediate steps such as tokenization, lemmatization, stemming, and dimensionality reduction.

\subsection{Named Entity Recognition}
Named Entity Recognition (NER) is a NLP task that involves identifying proper nouns or rigid entities in text such as organizations, persons, locations, and quantities \cite{li2020survey1, sun2018overview, yadav2019survey}. This is an important task for various applications like question answering, information retrieval, relation extraction, text summarization, and machine translation. There are four major categories of NER techniques, including Rule-based NER that uses handcrafted rules, unsupervised learning approaches that rely on clustering based on contextual similarity, supervised learning approaches that employ feature engineering with conventional classification schemes such as HMM, SVM, or other classifiers, and Deep learning-based approaches \cite{li2020survey1, sun2018overview, yadav2019survey}.

\subsection{Part-of-Speech tagging}
Part-of-speech (POS) tagging is an NLP task that involves assigning a specific category to each token in a given text, such as verb, noun, or adjective. POS tagging methods are classified into four categories: rule-based, stochastic, transformation-based, and HMM-based \cite{Jurafsky2000SpeechAL}. Rule-based taggers use a set of predefined rules to tag words. Stochastic methods, on the other hand, are probability-based and use frequency information to disambiguate the tags. Transformation-based taggers combine both rule-based and stochastic methods, where a small set of rules are learned from the data and tagging is transformed in each cycle. HMM-based taggers find the most probable sequence of POS tags for a given input sequence, modeled using Hidden Markov Model (HMM) \cite{Brants2000TnTA, Joshi1985NaturalLP}.

\subsection{Machine comprehension and Question answering}
Question answering is a task that has been extensively studied in the NLP literature, with the goal of building systems that can automatically generate answers to questions posed in a given context. This task has numerous applications, including the development of chatbots and dialogue generation systems. In order to train a system to perform this task, a neural network is trained on a large dataset of contexts, questions, and their respective answers, learning the relationships among them. The SQuAD dataset \cite{rajpurkar2016squad}, for example, has been proposed for this purpose, containing a large number of context paragraphs, questions, and answers.

\subsection{Automatic summarization}
Automatic text summarization is a challenging NLP task that involves creating a shorter version of a larger text document. The task requires the selection of essential information from the entire text and condensing it using the sentences available in the document or a different set of vocabulary. There are two ways to perform text summarization: extractive summarization and abstractive summarization, which are determined by the available training data and the desired output. Extractive summarization involves selecting keywords, phrases, and lines from the document and combining them to create a summary, making it useful when the available training data is limited \cite{DBLP:conf/nips/HermannKGEKSB15, roush-balaji-2020-debatesum}. On the other hand, abstractive summarization is a data-driven approach that involves training a machine learning model to create a summary using the learned language in the available data \cite{DBLP:conf/nips/HermannKGEKSB15, xsum}. Abstractive summarization requires a large amount of data but produces a high-quality summary of the text document.

\subsection{Machine translation}
The task of machine translation involves the conversion of text from a source language to a target language, which can be either in the form of sentences or documents. Based on the availability of a large training corpus and similarities between languages, MT systems can be modeled as either monolingual or multilingual \cite{yang2020survey, britz2017massive, ramesh2021samanantar}. However, the task becomes more complex when dealing with diverse domains such as legal, medical, and cultural texts. Additionally, the mode of text, whether formal written \cite{wmt-2021-machine} or informal spoken \cite{iwslt-2022-international}, poses a challenge for the task. Moreover, machine translation is susceptible to hallucinations \cite{DBLP:conf/naacl/RaunakMJ21} and can be vulnerable to adversarial attacks.

\subsection{Dialogue generation}
Automatic dialogue generation is a crucial NLP task that focuses on building conversational systems capable of generating responses automatically, with the goal of creating a human-like conversation system. These systems can be either task-oriented, designed to operate within a specific domain such as transportation, restaurants, or shopping, or open dialogue systems for open-ended conversations, such as everyday conversation between two people. Dialogue generation systems have evolved over three phases, including rule-based systems, retrieval-based systems , and neural generative conversation systems \cite{sun2021neural}. The quality of the generated response is dependent on the system's intelligence, as it should be consistent with the previous statement and align with the context of the conversation and targeted domain (if applicable).

\subsection{Question generation}
Automatic question generation is a NLP task that is designed to facilitate educational assessment by generating questions automatically from a given text. This task is important as it helps to reduce manual labor and time. The questions generated by these systems should be well-phrased and must have answers that can be found in the given piece of text. These questions can be of two types: subjective or objective \cite{das2021automatic}. Question generation systems have applications in a variety of fields such as Massive Open Online Courses (MOOC), healthcare systems, chatbots, and search engines \cite{nwafor2021automated}.

\section{Illustrations of some adversarial defense methods in NLP}
In this section some of the popular adversarial defense schemes in NLP are described with illustrations, across various categories proposed in the paper.

 \subsection{AdvEntuRe: Adversarial Training for Textual Entailment with Knowledge-Guided Examples}
\begin{figure*}[h!]
    \centering
    \includegraphics[scale=0.5]{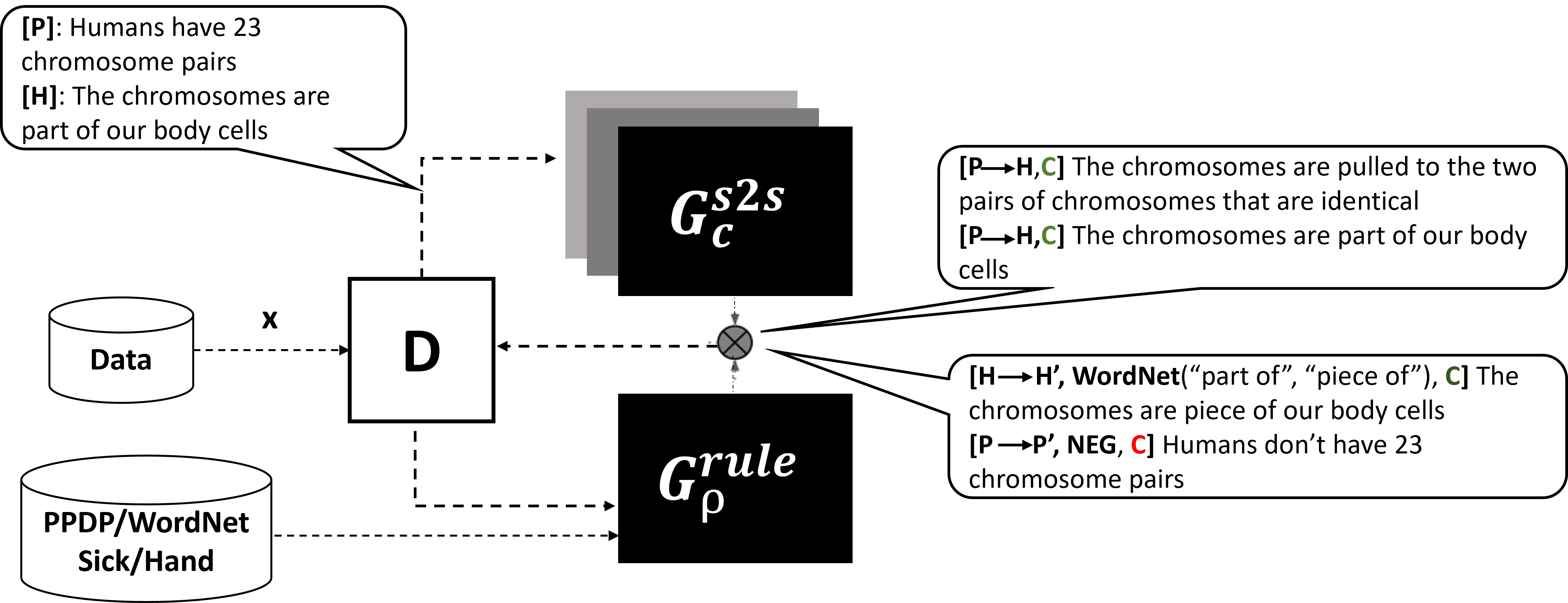}
    \caption{Overall pipeline of Adventure- Knowledge guided textual entailment \cite{kang2018adventure}}
    \label{fig:adventure}
\end{figure*}

  In the direction of GAN based adversarial training, the work proposed in "AdvEntuRe: Adversarial Training for Textual Entailment with Knowledge-Guided Examples" \cite{kang2018adventure} uses adversarial training with GANs for the textual entailment task. The generator and discriminator are trained in an end-to-end manner, where generator (seq2seq) is trained for generating adversarial examples using external knowledge or handwritten rules. Discriminator is trained in the same manner to learn the textual entailment for the generated samples. Figure \ref{fig:adventure} presents the overall pipeline proposed in this work for GAN-based adversarial training.

 \subsection{Natural Language Adversarial Defense through Synonym Encoding}

\begin{figure*}[h!]
    \centering
    \includegraphics[scale=0.5]{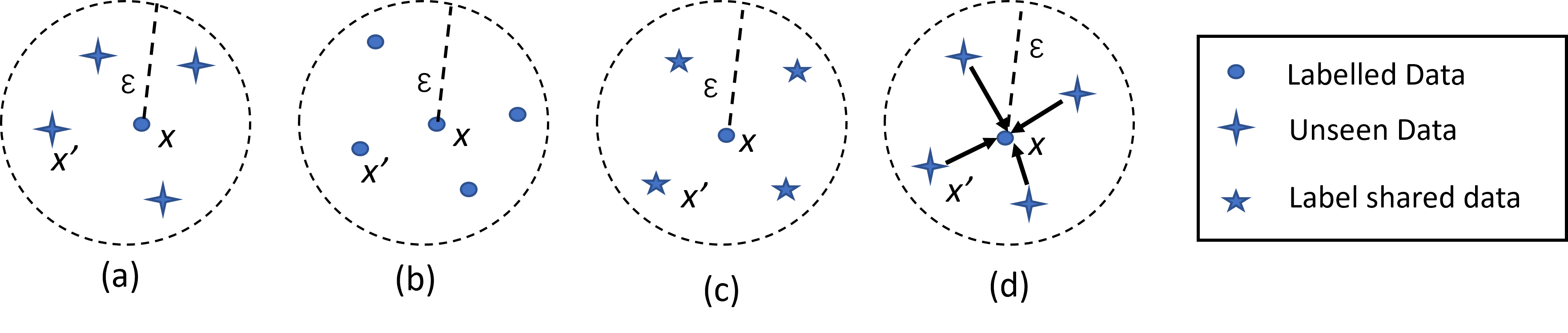}
    \caption{The neighborhood of a data point x in the input space in \cite{wangnatural}. (a) Normal training has unseen data points $x$, which leads to wrong classification. (b) Training with infinite labeled data to overcome all possible adversarial attacks. (c) Training with data with shared labels, in which all the neighboring points share the labels. (d) Mapping the neighboring points to center $x$ to remove adversaries. }
    \label{fig:word_level}
\end{figure*}

In the direction of perturbation control based adversarial defense methods, "Natural Language Adversarial Defense through Synonym Encoding"  \cite{wangnatural} proposed perturbation identification scheme. Perturbations related to word modifications which included insertion, deletion, substitution or swapping of words are identified in several ways. In this work, the authors proposed defense mechanism, against synonym substitution, calling it “Synonym Encoding Method”(SEM). They essentially clustered all the synonyms in embedding space with their euclidean distances and then encoder is layered before input to train the model. Encoder is responsible for identifying all the synonym substitution-based attacks in the model and maps all the synonyms to a unique encoding without adding extra data for training. Figure \ref{fig:word_level} demonstrates the Synonym Encoding Method proposed described above.  

\subsection{Interpretable Adversarial Perturbation in Input Embedding Space for Text}
\begin{figure*}[h!]
    \centering
    \includegraphics[scale= 0.5]{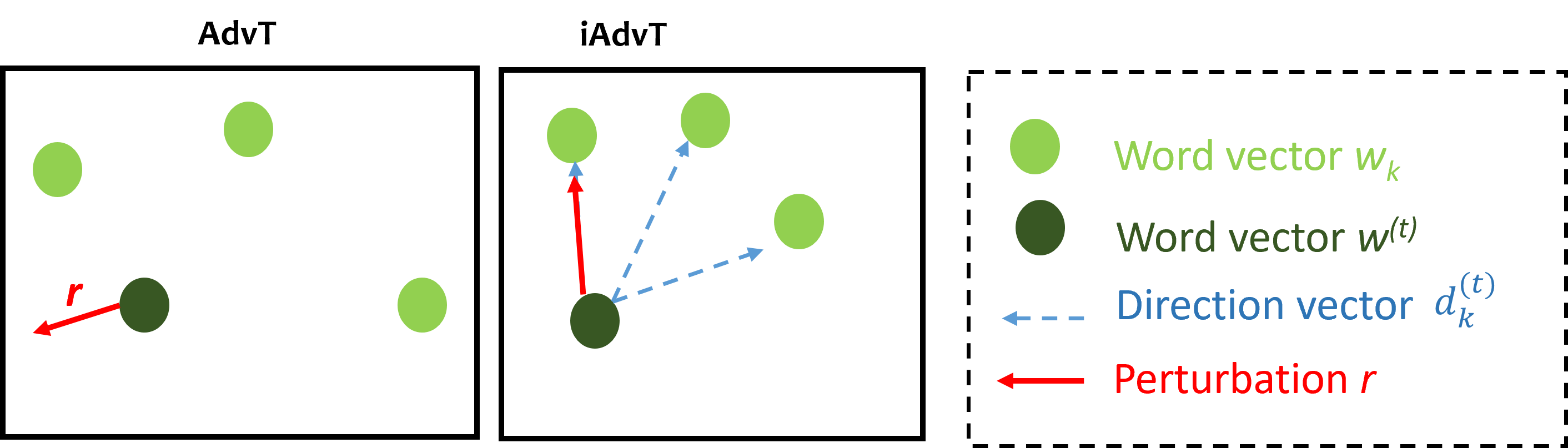}
    \caption{Restricting the direction of the perturbations proposed in \cite{sato2018interpretable}. The proposed method (iAdvT) restricts the perturbations in the direction of input word embeddings, while the previous methods (AdvT) lets them choose any direction}
    \label{fig:advt}
\end{figure*}

The proposed work  "Interpretable Adversarial Perturbation in Input Embedding Space for Text" \cite{sato2018interpretable}, under perturbation direction control category alters the direction of the perturbations towards the cleaner text input limiting the adversarial space. Along this line, this work proposed an interpretable adversarial training method by restricting the direction of adversarial samples. The direction of perturbation is restricted to the words in the existing vocabulary so that perturbations could be interpreted even after adversarial training. Figure \ref{fig:advt} shows the mechanism to restrict the direction of the perturbations.

\subsection{SAFER: A Structure-free Approach for Certified Robustness to Adversarial Word Substitutions}

\begin{figure*}[h!]
    \centering
    \includegraphics[scale=0.5]{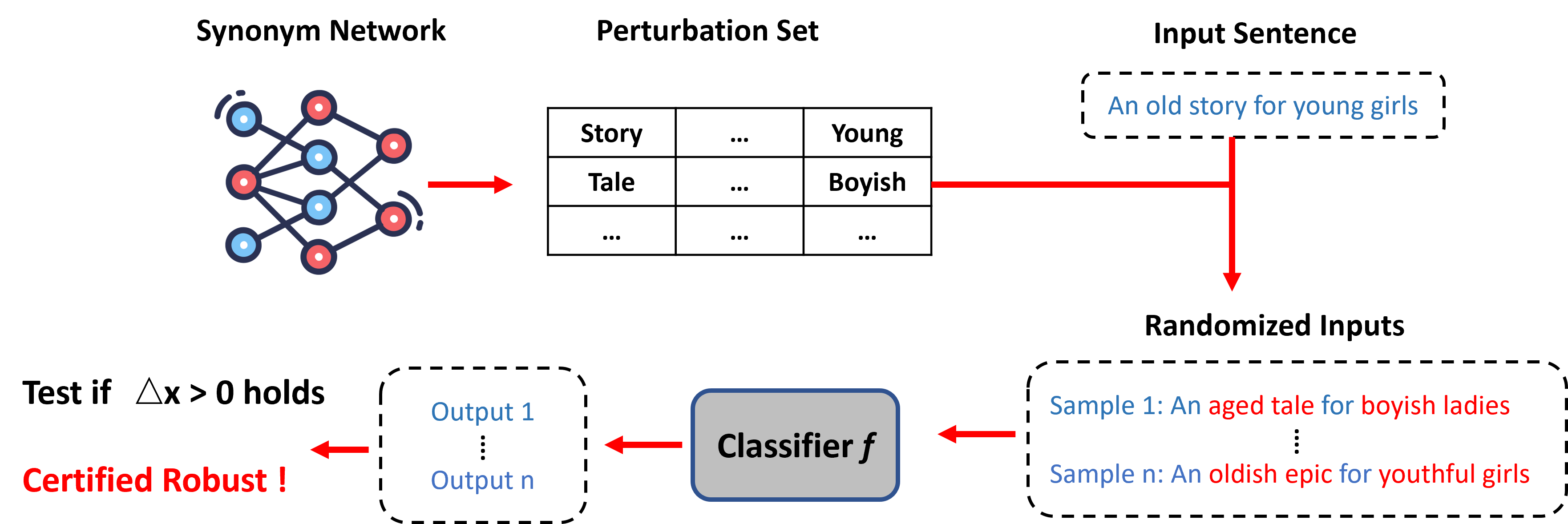}
    \caption{Pipeline for certified robustness approach proposed in \cite{ye2020safer}}
    \label{fig:safer}
\end{figure*}

In the work, "SAFER: A Structure-free Approach for Certified Robustness to Adversarial Word Substitutions" \cite{ye2020safer} in the direction of robustness by certification, proposed structure-free certified robust models which can be applied to any arbitrary model. This method overcomes the limitations of IBP based method in which they are not applicable to character level and sub-word level models. They prepared a perturbation set of words using synonym sets, top-K nearest neighbors under the cosine similarity of GLOVE vectors, where K is a hyperparameter that controls the size of the perturbation set. They further generated sentence perturbations using word perturbations and trained a classifier with robust certification. 
Figure \ref{fig:safer} presents the overall pipeline to achieve certified robustness. In the context of IBP methods, 

\subsection{Certified Robustness to Word Substitution Attack with Differential Privacy}
\begin{figure*}[h]
    \centering
    \includegraphics[scale=0.5]{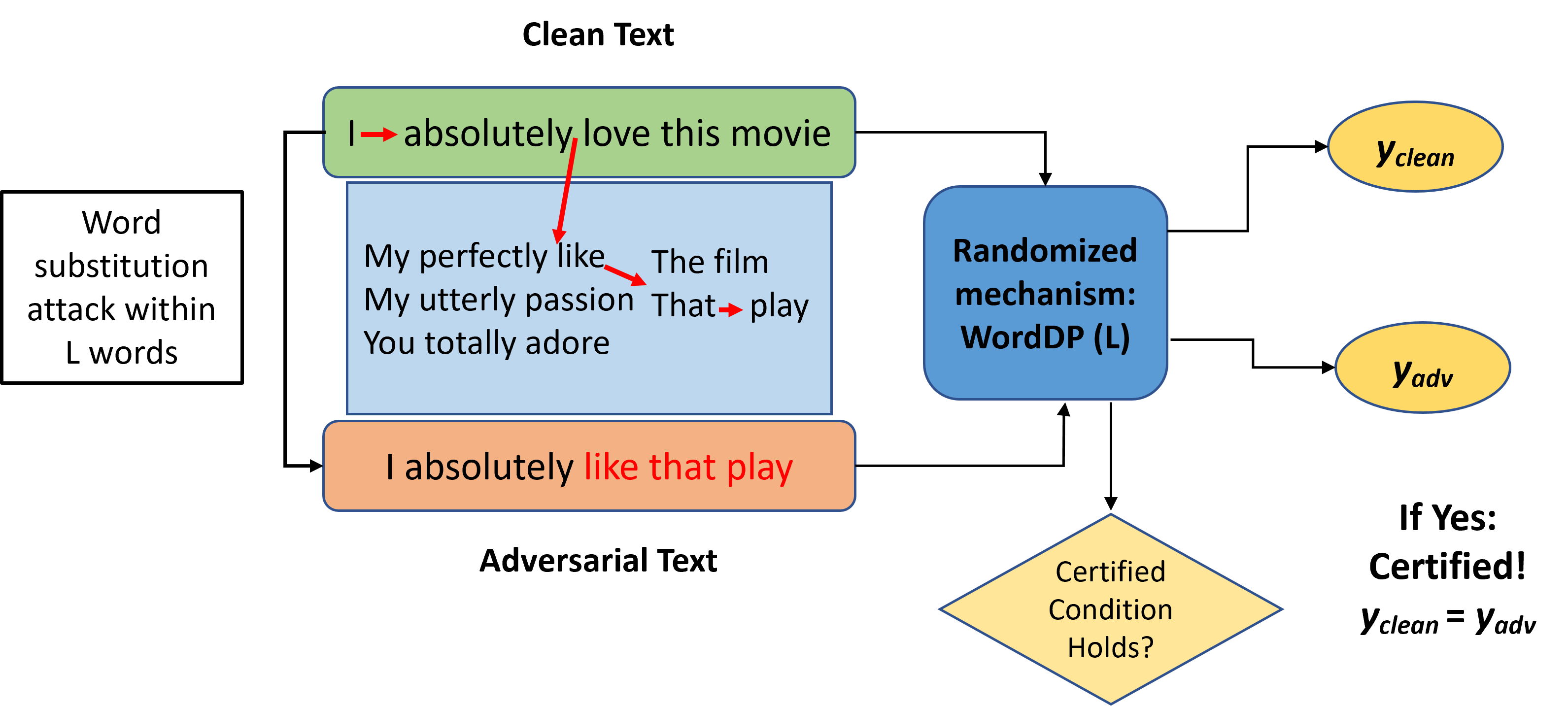}
    \caption{Word substitution attack and certified robustness via wordDP proposed in \cite{wang2021certified} }
    \label{fig:worddp}
\end{figure*}

The work "Certified Robustness to Word Substitution Attack with Differential Privacy" \cite{wang2021certified} proposed in the direction of robustness by certification,  provided a certificate of robustness with the idea of differential privacy in the input data. They implemented differential privacy in the textual data by treating a sentence as a database and words as an individual records. If a predictive model satisfies a certain threshold (epsilon-DP) for a perturbed input, its input should be the same as the clean data. Hence providing a certification of robustness against L-adversary word substitution attacks. 
Figure \ref{fig:worddp} demonstrates the certified robustness framework proposed in \cite{wang2021certified}.

\subsection{Adversarial NLI: A New Benchmark for Natural Language Understanding }
\begin{figure*}[h]
    \centering
    \includegraphics[scale=0.5]{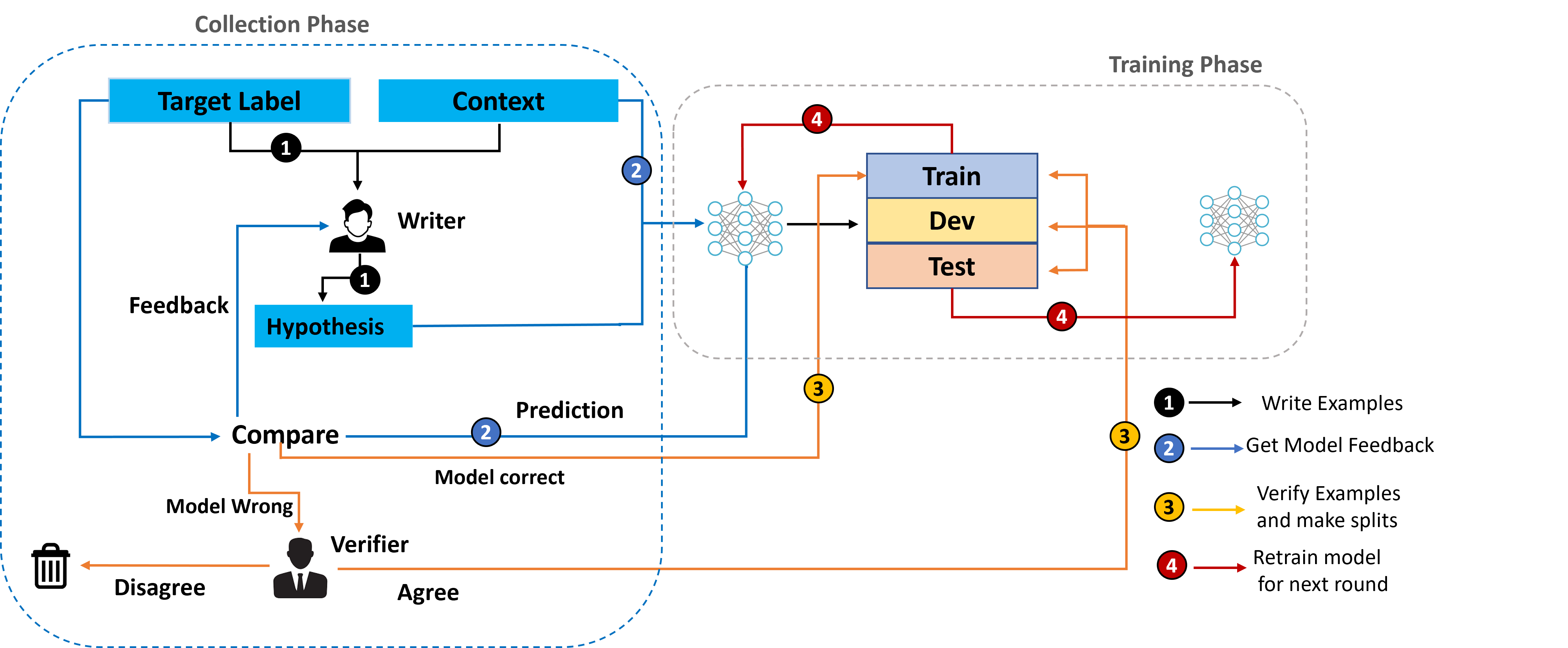}
    \caption{Adversarial NLI data collection framework with Human-and-model-in-the-Loop proposed in \cite{nie2019adversarial}}
    \label{fig:anli}
\end{figure*}

In the direction of adversarial datasets and framework, ANLI \cite{nie2019adversarial}, proposed adversarial dataset for natural language inference systems. ANLI composed of adversarial examples for NLI collected in 3 iterative rounds having human and machine in loop.  ANLI consist of  $103k$ examples of sentences,  starting with short multi-sentence passages from Wikipedia and having annotators writing adversarial hypothesis. In this process, they tested these samples with state-of-the-art NLI models and got then verified by human annotators hence proposing human-and-model-in-the-loop enabled training (HAMLET) scheme for data collection. The adversarial examples collected in this dataset are used in adversarial training to improve the performance of Natural Language Inferencing models. 
Figure \ref{fig:anli} depicts the dataset collection framework for ANLI dataset proposed in \cite{nie2019adversarial}. 

\bibliographystyle{ACM-Reference-Format}

\bibliography{acmart}
\end{document}